\documentclass[smallcondensed
]{svjour3}     % onecolumn (ditto)
\smartqed  % flush right qed marks, e.g. at end of proof
\usepackage{graphicx}
\usepackage{epstopdf}
\usepackage{geometry}
\usepackage{comment}
\usepackage{xspace}
\usepackage{amsmath}
\usepackage{amssymb}

\usepackage{epsfig}
\usepackage{subfigure}
\usepackage{algorithmic}
\usepackage{graphics}
\usepackage{alltt}
\usepackage{float}
\usepackage{array}
\usepackage{footmisc}
\usepackage{comment}
\usepackage{algorithm}

\usepackage[numbers]{natbib}
\usepackage{bm}
\usepackage{multirow}

\usepackage{epsfig}
\usepackage{wrapfig}
\usepackage{url}
\usepackage{amsmath,amssymb,amsfonts}
\usepackage{graphicx}
\usepackage{textcomp}

\usepackage[justification=centering]{caption}
\usepackage[misc]{ifsym}

\newtheorem{Definition}{Definition}

\begin{document}

\title{PAC-GAN: An Effective Pose Augmentation Scheme for Unsupervised Cross-View Person Re-identification%\thanks{Grants or other notes
}


\author{Chengyuan Zhang$^\dagger$ \and Lei Zhu$^\dagger$ \and ShiChao Zhang$^\dagger$ %\and Weiren Yu$^\ddagger$
}


\institute{Chengyuan Zhang \at
              \email{cyzhang@csu.edu.cn}           %  \\
           \and
           \Letter Lei Zhu \at
              \email{leizhu@csu.edu.cn}           %  \\
           \and
           \Letter ShiChao Zhang \at
              \email{zhangsc@csu.edu.cn}
              \and
%            Weiren Yu \at
%              \email{w.yu3@aston.ac.uk}           %  \\
              \and
           $^\dagger$ School of Computer Science and Engineering, Central South University, Changsha, 410083, China\\
%$^{\ddagger}$ School of Computer, Hunan University of Technology, China\\
}

\date{Received: date / Accepted: date}

\maketitle

\begin{abstract}
Person re-identification (person Re-Id) aims to retrieve the pedestrian images of a same person that captured by disjoint and non-overlapping cameras. Lots of researchers recently focuse on this hot issue and propose deep learning based methods to enhance the recognition rate in a supervised or unsupervised manner. However, two limitations that cannot be ignored: firstly, compared with other image retrieval benchmarks, the size of existing person Re-Id datasets are far from meeting the requirement, which cannot provide sufficient pedestrian samples for the training of deep model; secondly, the samples in existing datasets do not have sufficient human motions or postures coverage to provide more priori knowledges for learning. In this paper, we introduce a novel unsupervised pose augmentation cross-view person Re-Id scheme called PAC-GAN to overcome these limitations. We firstly present the formal definition of cross-view pose augmentation and then propose the framework of PAC-GAN that is a novel conditional generative adversarial network (CGAN) based approach to improve the performance of unsupervised corss-view person Re-Id. Specifically, The pose generation model in PAC-GAN called CPG-Net is to generate enough quantity of pose-rich samples from original image and skeleton samples. The pose augmentation dataset is produced by combining the synthesized pose-rich samples with the original samples, which is fed into the corss-view person Re-Id model named Cross-GAN. Besides, we use weight-sharing strategy in the CPG-Net to improve the quality of new generated samples. To the best of our knowledge, we are the first try to enhance the unsupervised cross-view person Re-Id by pose augmentation, and the results of extensive experiments show that the proposed scheme can combat the state-of-the-arts.

\keywords{Cross-view Person Re-Id \and Pose Agumentation \and Generative Adversarial Networks \and Unsupervised Learning}

\end{abstract}

\section{Introduction}
\label{sec:introduction}
\begin{figure}
  %\newskip\subfigtoppskip \subfigtopskip = -0.1cm
  \centering
  \includegraphics[width=1.0\linewidth]{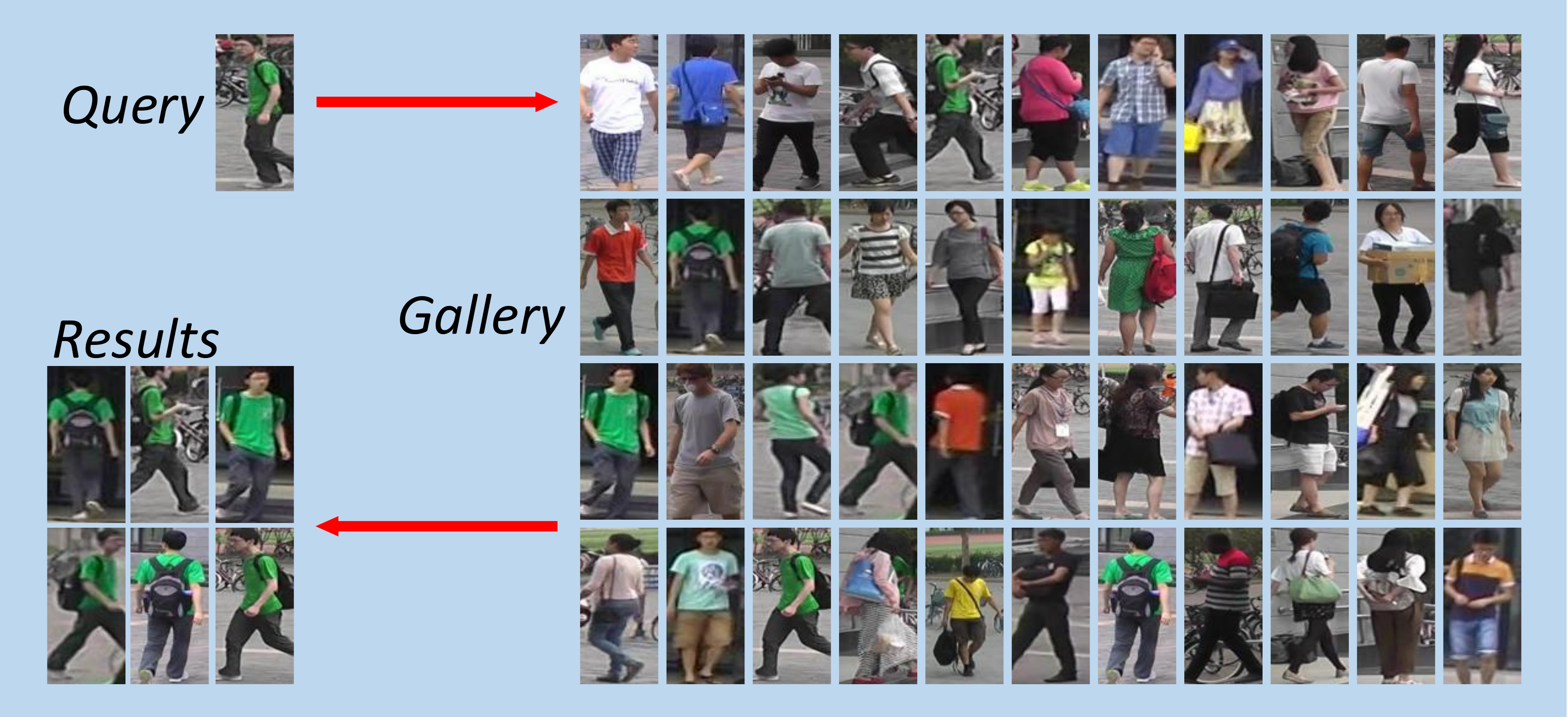}
  \vspace{-1mm}
  \caption{\small An example of person Re-Id task. The samples shown in this figure are chosen from Market-1501~\cite{DBLP:conf/iccv/ZhengSTWWT15}.}
  \label{fig:person-re-id}
\end{figure}

In recent years, person re-identification (person Re-Id) problem has attracted attention of lots of researchers with the wide application of video surveillance system. This problem is a specific computer vision task that is to retrieve the target images from the gallery by a query that contains the same pedestrian in the targets through spatially disjoint camera views, as shwon in Fig~\ref{fig:person-re-id}. Cross-view person Re-Id aims to make up for the visual limitations of the current location-fixed cameras, and can be combined with pedestrian detection/tracking techniques. It can be widely used in many applications such as intelligent video surveillance, intelligent security and forensic search.

\begin{figure*}
  %\newskip\subfigtoppskip \subfigtopskip = -0.1cm
  \centering
  \includegraphics[width=1.0\linewidth]{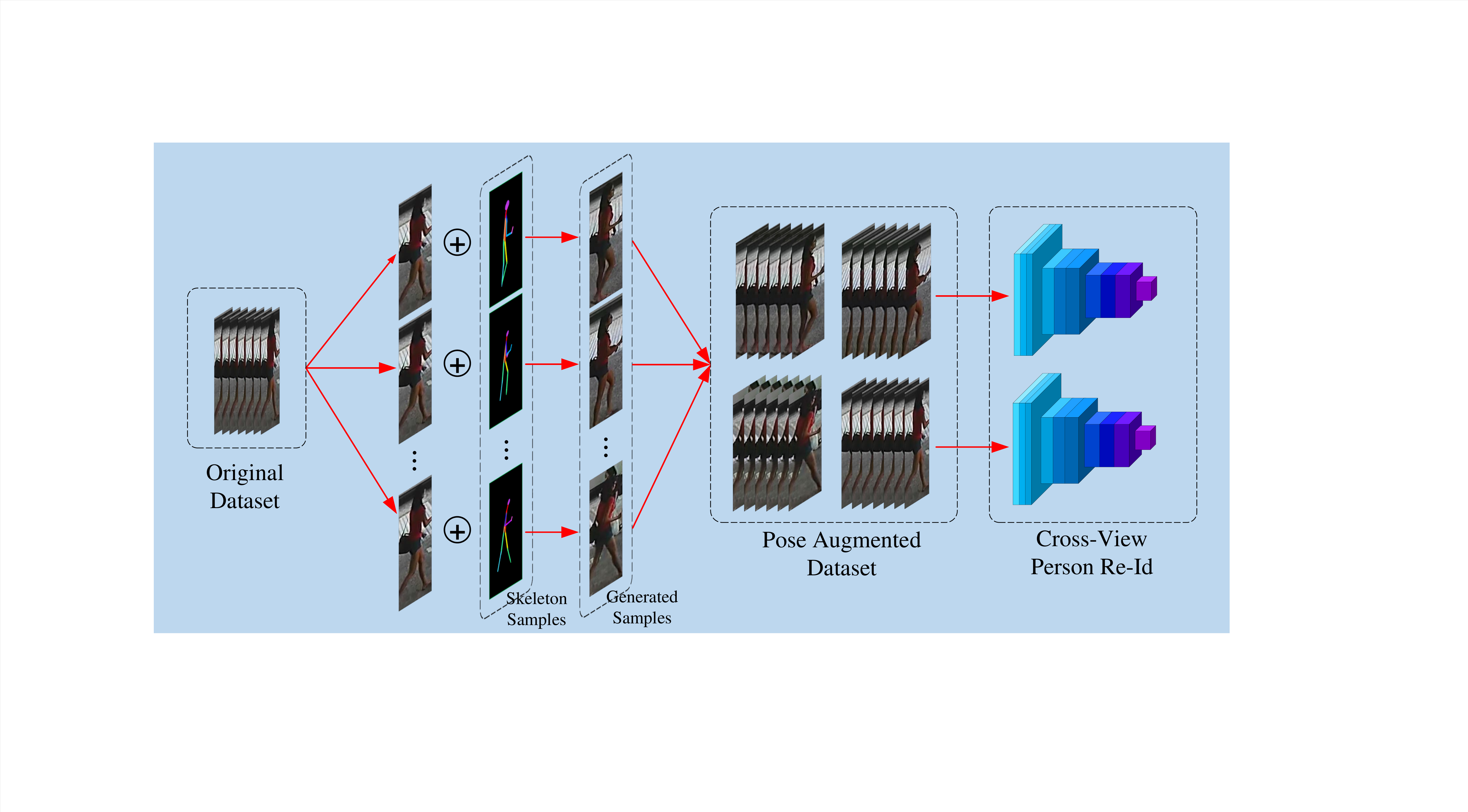}
  \vspace{-1mm}
  \caption{\small  Pose Augmentation for Person Re-identification. The original person Re-Id dataset has ordinarily not enough sufficient poses coverage, which is a limitation for person Re-Id task. To overcome this challenge, we propose to generate a pose-rich dataset that contains enough quantity of new samples by using skeleton samples and original pedestrian samples. The pose augmented dataset that includes original samples and new produced pose-rich samples can provide more priors for cross-view person Re-Id training.}
  \label{fig:motivation}
\end{figure*}

Deep learning techniques~\cite{DBLP:journals/nature/LeCunBH15,DBLP:journals/nn/Schmidhuber15}, especially deep convolutional neural networks~\cite{Krizhevsky2012ImageNet} (CNNs), show a great superiority on the tasks of computer vision and image retrieval. A growing number of Deep learning based methods are proposed to address cross-view person Re-Id in two main aspects: feature extraction and metric learning. he first category of methods~\cite{Chen2016Similarity,YangyangECCV2014,Rama2016Learning,Wu2016An,Ahmed_2015_CVPR,Varior2016Gated,Lin2017Deep} aims to generate effective discriminative representations via learning common or relevant visual features from cross-view samples to combat the view variations. The other category of approaches~\cite{Chen_2017_ICCV,Liao2015Person,Lin20193Deep,Lin2019entropy,Lin2019TCYB,Zhou2017Efficient,Bak_2017_CVPR,Zhang_2016_CVPR,Li_2013_CVPR} employ a variety of different hand-crafted visual features such as color histogram, local maximal occurrence and local binary patterns to learn a similarity metric to measure the visual similarity between samples. Lately more deep learning based approaches are desinged as a unified framework that consists of both feature learning method and metric learning technique, which is to extract deep visual features from cross-view samples by coupled CNNs and a metric learning module measure the similarity of inputs.

\textbf{Motivation.} Although great progress has been made in cross-view person Re-Id research, there are still two undeniable limitations. The first is the existing person Re-Id datasets such as VIPeR~\cite{Gray2007Evaluating}, CUHK03~\cite{Li_2014_CVPR}, Market-1501~\cite{DBLP:conf/iccv/ZhengSTWWT15} are far from large enough compared with other image retrieval benchmarks, which cannot provide sufficient pedestrian samples for the training of deep model. The second is that the samples in existing datasets do not have abundant human motions or poses coverage to provide more priori knowledges for learning. Even through MARS~\cite{DBLP:conf/eccv/ZhengBSWSWT16}, is an extension of the Market-1501, has been proposed for human motion analysis and person Re-Id, the number of pedestrian identities in this benchmark is still not large enough. Zheng et al~\cite{Zheng_2017_ICCV} utilized DCGAN~\cite{Radford2015Unsupervised} to produce more pedestrian samples to improve the discriminative ability, but they just focused on the number of samples, rather than the richness of human poses in datasets.

To combat these limitations, this work proposes to extend existing person Re-Id datasets by developing a novel unsupervised data augmentation approach that aims to generate enough quantity of pose-rich pedestrian samples. As illustrated in Fig.~\ref{fig:motivation}, the generated samples are produced according to original pedestrian samples that provide foreground information of person appearance and skeleton samples that provide the sufficient poses information. These synthesized pose-rich samples are combined with these benchmarks can provide much more priors for training. Integrated with an unsupervised cross-view person Re-Id model, a novel unsupervised pose augmentation person Re-Id framework is created. To the best of our knowledge, we are the first to enhance the performance of unsupervised cross-view person Re-Id by pose augmentation.

\textbf{Our Approaches.} To implement the idea aforementioned, we propose a novel unsupervised pose augmentation person Re-Id framework named \textbf{PAC-GAN} based on generative adversarial networks (GAN). This framework consists of two main models: the one is called \textbf{C}ross-view \textbf{P}ose \textbf{G}eneration \textbf{Net} (\textbf{CPG-Net}) that is a conditional GAN based network with a coupled structure to generate pose-rich samples to augment the person Re-Id dataset in the aspect of human motions or postures. This model includes two branches that named PG-Net-$V_1$ and PG-Net-$V_2$ for two different camera views respectively. Each of them receives paired inputs of skeleton samples  and pedestrian appearance samples. To further improve the quality of new produced cross-view samples, we apply weight-sharing strategy across the coupled generative networks and discriminative networks in CPG-Net to capture the co-occurrence visual patterns of original cross-view images. The generated pose-rich samples is combined with the original cross-view person Re-Id dataset as the pose augmented dataset to support unsupervised person Re-Id.

The other part of PAC-GAN is an effective unsupervised cross-view person Re-Id model named \textbf{Cross-GAN} presented in our privous work. A coupled variational auto-encoder (VAE) with a cross-view alignment is used to map the cross-view images into latent variables, and a coupled GAN layer receives the corss-view codes to learn the joint distribution of cross-view samples.

For the skeleton samples generation, two skeleton generation modules with the same structure are employed in this framework to generate the skeleton samples from image datasets that have a wide range of human poses coverage, one per camera view. To achieve better generation performance, we utilize the realtime human pose estimator method introduced by~\cite{Cao_2017_CVPR}, which is the state-of-the-art method of human pose estimation.

\textbf{Contributions.} The contributions of this work are three-fold:
\begin{itemize}
    \item We introduce the definition of the cross-view pose augmentation for person Re-Id task in formal and propose a novel scheme of pose augmentation for unsupervised cross-view person Re-Id. To the best of our knowledge, this work is the first time to improve the performance of unsupervised corss-view person Re-Id by pose augmentation.
    \item We propose a novel cross-view pose generation model named CPG-Net. This Conditional GAN based model can generate new samples that have various poses and the same visual appearance from skeleton and pedestrian samples. Besides, we proposed to apply weight-sharing strategy in the generator and discriminator of CPG-Net to learn the co-occurrence visual patterns from inputs, which can effectively improve the quality of sample generation.
    \item We compare our approach with semi/un-supervised and supervised state-of-the-arts on VIPeR, CUHK03 and Market-1501 benchmark datasets. The results show that our approach can improve the recognition rate effectively.
\end{itemize}

\textbf{Roadmap.} The remainder of this paper is organized as follows:  We review the related works in Section~\ref{sec:relate-work}. Section~\ref{sec:preliminaries} introduces the definition of person Re-Id and cross-view pose augmentation, as well as the basic theories and techniques involved in this paper. We propose a novel cross-view pose augmentation person Re-Id framework named PAC-GAN in section~\ref{sec:Methodology}. Section~\ref{sec:experiments} discusses the externsive experiments of the proposed approach and the state-of-the-arts, and finally we draw our conclusion of this paper in Section~\ref{sec:conclusion}.

\section{Related Works}\label{sec:relate-work}
In this section, we introduce an overview of previous studies of person Re-Id, generative adversarial networks and pose estimation, which are related to this work. To the best of our knowledge, there is no existing work to improve the accuracy of cross-view person Re-Id in an unsupervised or semi-supervised manner by using a pose augmentation process for each camera view.

\subsection{Person Re-identification}
Person Re-Id~\cite{DBLP:journals/pr/WuWLG18,Shen_2015_ICCV,DBLP:books/daglib/0034533,DBLP:conf/cvpr/FarenzenaBPMC10} is a hot issue in the field of visual recognition and image processing. It is a task to associate and match pedestrians across camera views at different geo-locations and times in a distributed multi-cameras surveillance system~\cite{DBLP:books/daglib/0034533}. With the widespread application of multi-cameras networks and video surveillance systems, lots of researchers paid more attations on this problem in the past few years by using supervised learning, unsupervised learning and semi-supervised learning techniques.

\subsubsection{Person Re-Id via Supervised learning}
Supervised learning techniqes are the most commonly used to solve the person Re-Id problem, which are adopt by lots of researchers in the aspects of invariant feature learning and metric learning~\cite{DBLP:journals/ftml/Kulis13} to deal with this challenge. For feature extraction and learning, Gray et al.~\cite{DBLP:conf/eccv/GrayT08} proposed to utilize the ensemble of localized features (ELF), an efficient object representation to realize viewpoint invariant pedestrian recognition. To overcome the adversarial effect by pose variation, Chen et al.~\cite{Chen2016Similarity} proposed a novel similarity framework consisting of multiple sub-similarity measurements, which is based upon polynomial feature map to describe the matching within each subregion. This framework can collabrate both local and gobal similarity to exploit their complementary strength. Yang et al.\cite{YangyangECCV2014} introduced a novel salient color names based color descriptor called SCNCD for person Re-Id task. They formulated the person Re-Id task as a color distribution matching problem, and the effect of background information is utilized to improve the accuracy of recognition. To handle the problem of lighting condition changes across different camera views, Rahul Rama Varior et al.~\cite{Rama2016Learning} proposed a novel framework for learning color patterns from across view. In this scheme, color feature generation is modeled as a learning problem by jointly learning a linear transformation and a dictionary to encode pixel values. Nanda et al.~\cite{DBLP:journals/mta/NandaCSB19} presented a novel multi-shot person Re-Id framework to solve the illumination variations problem by a images preprocessing step. In order to improve the performance of pedestrian discrimination, Zheng et al.~\cite{DBLP:journals/tomccap/ZhengZY18} proposed to combine verification and identification models to generate more discriminative pedestrian descriptors. A siamese network architecture was designed, which can recognise the two inputs which are belong to the same identity or not.  Zhao et al.~\cite{Zhao_2014_CVPR} proposed to generate mid-level filters from automatically discovered patch clusters for person Re-Id task, which can identify specific visual patterns and have fine cross-view invariance.

Recently, Deep learning~\cite{DBLP:journals/pr/WuWLG18,DBLP:journals/nature/LeCunBH15,WangTIP17,WuTIP19,WuTIP192,WangIJCAI16,WangTIP15,WangTNNLS18,DBLP:journals/nn/Schmidhuber15,DBLP:journals/ijon/GuoLOLWL16,DBLP:journals/corr/RadfordMC15,DBLP:journals/cviu/WuWGHL18,Where-and-WhenTMM} technique as a kind of exceedingly powerful and efficient tool is applied for the tasks of person Re-Id. Combining hand-crafted histogram features and CNN features, Wu et al.~\cite{Wu2016An} presented a novel feature extraction model called Feature Fusion Net (FFN) to generate a novel deep feature representation which is more discriminative and compact. Based on deep convolutional neural networks, Chen et al.~\cite{DBLP:journals/tip/ChenGL16} formulated person Re-Id as a learning-to-rank problem and introduced a unified deep ranking framework to learns a similarity metric. Chen et al.~\cite{Chen_2017_ICCV} proposed to jointly learn discriminative scale-specific features and maximise multi-scale feature fusion selections to improve the performance of discrimination. They designed a novel Deep Pyramid Feature Learning (DPFL) CNN architecture to fuse multi-scale appearance features. Ahmed et al.~\cite{Ahmed_2015_CVPR} presented a new deep neural network architecture that formulates the problem of person re-identification as binary classification. This model has two novel layers, one is a cross-input neighborhood differences layer, and the other is a subsequent layer that summarizes these differences. Wu et al.~\cite{Where-and-WhenTMM} introduced a novel deep siamese architecture that jointly learns spatio-temporal video representations and similarity metrics. They used attention mechanism to select the most relevant features during the recurrence to attend at distinct regions in cross-view. In another work of them~\cite{DBLP:journals/cviu/WuWGHL18}, a deep hashing framework with Convolutional Neural Networks (CNNs) for fast person re-identification was developed. By this scheme, both CNN features and hash functions are simultaneously learned to get robust yet discriminative features and similarity-preserving hash codes.

The supervised learning frameworks above-mentioned have to rely on labelled images to generate discriminative visual features. However, in practical environment, it is very expensive to annotate images in a large-scale video surveillance system, whcih is a limitation for the scalability of their application.

\subsubsection{Person Re-Id via Unsupervised learning}
To directly utilize unlabeled data for person Re-Id task, Unsupervised learning~\cite{Bengio2012Unsupervised,DBLP:journals/tip/ChenGL16,DBLP:conf/eccv/WeberWP00} has been used as a reasonable and ingenious technique in many studies. Ma et al.~\cite{DBLP:conf/eccv/MaSJ12} proposed a novel Fisher Vectors based descriptor to solve person Re-Id problem in an unsupervised manner. Liu et al.~\cite{DBLP:conf/eccv/LiuGLL12} introduced a novel unsupervised method for learning a bottom-up feature importance, which is based on the idea that under different circumstances certain visual features are more important than others for distinguishing one person from others. Inspired by the idea that some small salient visual information can be used to discriminate different person, Zhao et al.~\cite{Zhao_2013_CVPR} proposed to a novel framework to learn human salience in an unsupervised way to address person Re-Id problem. Farenzena et al.~\cite{DBLP:conf/cvpr/FarenzenaBPMC10} developed an appearance-based method which extracts visual features from three complementary aspects of the pedestrian appearance. This scheme is robust to low resolution, occlusions and pose, changes of viewpoint and illumination. Based on probabilistic generative theme modeling, Wang et al.~\cite{DBLP:conf/bmvc/WangGX14} presneted a novel unsupervised modeling approach to saliency detection, which can discover localised person foreground appearance saliency and remove busy background clutter surrounding a person simultaneously. Liang et al.~\cite{DBLP:conf/mm/LiangHHZJX15} adopt probabilistic model to organize and depict the spatial feature distribution of person images, which is a robust approach against  environment changes and external interference. Ma et al.~\cite{DBLP:journals/pr/MaZGXHLZ17} proposed a novel video based person Re-Id approach to match pedestrian across views. A novel space-time person representation in form of sequence is generated based on existing action space-time features and spatio-temporal pyramids. Wang et al.~\cite{DBLP:conf/icip/WangZXG16} introduced a novel person Re-Id setting in an unsupervised manner named OneShot-OpenSet-ReID, and proposed an unsupervised subspace learning model named RKSL that can learn cross-view identity discriminative information from unlabeled data. Different from the studies above-mentioned, Peng et al.~\cite{Peng_2016_CVPR} proposed a novel cross-dataset unsupervised method named UMDL without any labelled matching pairs of target data by using cross-dataset transfer learning. To realize this model, they developed a new asymmetric multi-task learning approach that transfer a view-invariant representation from existing labelled datasets.

The other crucial fundamental problem, namely metric learning, is studied by many other works by using unsupervised techniques. Liao et al.~\cite{Liao2015Person} presented a subspace and metric learning method named Cross-view Quadratic Discriminant Analysis (XQDA). In this solution, a discriminant low dimensional subspace is learned by XQDA and simultaneously, a QDA metric is learned on the derived subspace. Yu et al.~\cite{Yu_2017_ICCV} proposed an unsupervised asymmetric metric learning model to learn specific projection for each view based on asymmetric clustering for cross-view person Re-Id. Zhou et al.~\cite{Zhou2017Efficient} proposed to shift part of the metric learning to online local metric adaptation, which only uses negative data from a negative sample database. Besides, this approach can achieve an adaptive nonlinear metric by combining a global metric with local metric adaptation. Bak et al.~\cite{Bak_2017_CVPR} presented a novel one-shot learning method to learn a metric containing texture and color components in an unsupervised manner. They address the problem of color differences  across camera views by using a single pair of ColorChecker images to learn a color metric.

Different from the aforementioned works, our previous work~\cite{DBLP:journals/ijon/ZhangWW19} proposed a novel model called crossing Generative Adversarial Network (Cross-GAN) for learning a joint distribution for cross-image representations in an unsupervised manner. As the practical configurations of pedestrian images are multi-modal and view-specific  even if they are observed under the same camera, we proposed to integrate variational auto-encoder with a cross-view alignment process into our model to encode the image pair into respective latent variables, which can reduce the view differences effectively. Besides, we used cross GAN with weight-sharing rather than the siamese convolutional neural networks (siamese CNNs)~\cite{Varior2016Gated} becasue they are composed of fixed receptive fields which may not flexible to capture the various local patterns.

\subsubsection{Person Re-Id via Semi-supervised learning}
Many semi-supervised learning~\cite{Zhu_Semi_Survey,DBLP:series/synthesis/2009Zhu} based studies for person Re-Id challenge have emerged in recent years. Figueira et al.~\cite{DBLP:conf/avss/FigueiraBMCBM13} proposed a novel solution consisting of a semi-supervised multi-feature learning strategy. This solution exploits multiple features independently and does not require training a classifier for each pair of cameras. To combat the issue of the variations in human appearances from different camera views, Liu et al.~\cite{Liu_2014_CVPR} proposed an efficient semi-supervised coupled dictionary learning method that requires only a small number of labeled images to carry the relationship between appearance features from different cameras. Abundant unlabeled training images are used to exploit the geometry of the marginal distribution. Ma et al.~\cite{DBLP:conf/accv/MaL14} manifested that positive prior in the rank-one matching subset is much larger than that in all the unlabeled data. According to this idea, a novel semi-supervised ranking approach was developed to utilize unlabeled data to improve the discrimination performance. In order to break the limitation of labeled image pairs available for training, Chen et al.~\cite{DBLP:conf/ACISicis/ChenCRLY17} presented a new semi-supervised KISS metric learning method. Zhu et al.~\cite{DBLP:journals/tcsv/ZhuJYYCGW18} introudced a novel semi-supervised cross-view projection-based dictionary learning (SCPDL) method for video person Re-Id task.

The studies aforementioned in both aspects of feature learning and metric learning are devoted to improve the performance of person Re-Id. However, the poses of the person in the existing benchmarks, e.g., VIPeR~\cite{Gray2007Evaluating}, CUHK03~\cite{Li_2014_CVPR}, Market-1501~\cite{DBLP:conf/iccv/ZhengSTWWT15}, etc. are not rich enough compared with the actual situation. Moreover, to the best of our knowledge, no attempt has been made to boost the accuracy of unsupervised person Re-Id via data augmentation process that makes the poses or motions of pedestrians more multifarious. In this paper, we propose a novel framework named PAC-GAN to effectively enhance the recognizing ability, which consists of two parts: (1) a new GAN based generative model named cross-view pose generation net (CPG-Net) and (2) Cross-GAN with weight sharing. The former is to generate pose-augmented visual data for two different camera views and the latter is to discriminate whether the persons in cross-view images are the same.

\subsection{Generative Adversarial Networks}
Generative Adversarial Networks (GAN for short) proposed by Goodfellow et al.~\cite{DBLP:conf/nips/GoodfellowPMXWOCB14} is one of the most prominent deep generative model. This powerful technique and its varieties such as CoGAN~\cite{DBLP:journals/nn/KiasariML18}, Triple-GAN~\cite{DBLP:conf/nips/LiXZZ17}, AdaGAN~\cite{DBLP:conf/nips/TolstikhinGBSS17}, AL-CGAN~\cite{DBLP:journals/corr/KaracanAEE16}, CGAN~\cite{DBLP:journals/corr/MirzaO14}, BiGAN~\cite{DBLP:conf/iclr/DonahueKD17}, CycleGAN~\cite{Zhu_2017_ICCV}, DCGAN~\cite{DBLP:journals/corr/RadfordMC15}, etc. are often utilized to solve wide variety of computer vision and pattern recognition problems, such as image-to-image translation and person Re-Id.

For the task of image-to-image translation that has attracted a lot of attention, Isola et al.~\cite{Isola_2017_CVPR} used conditional adversarial networks as a general-purpose approach for image-to-image translation tasks. In their work~\cite{Zhu_2017_ICCV}, they proposedan approach to translate an image from a source domain to a target domain without paired images. Yi et al.~\cite{Yi_2017_ICCV} presented a novel unsupervised learning approach called dual-GAN to train image translators from two sets of unlabeled images from two domains. Liu et al.~\cite{DBLP:conf/icpr/LiuGCL18} made an intensive study of CycleGAN~\cite{Zhu_2017_ICCV} and introduced two novel models named Long CycleGAN and Nest CycleGAN respectively to address the image-to-image translation problem. For the image-to-image translation task in a multi-modal scenario, Cherian et al.~\cite{DBLP:conf/wacv/CherianS19} proposed a semantically-consistent GAN framework called Sem-GAN with a segmentation module.

For the task of person Re-Id, Zheng et al.~\cite{Zheng_2017_ICCV} proposed to utilize deep convolutional generative adversarial network (DCGAN)~\cite{DBLP:journals/corr/RadfordMC15} to generate unlabeled samples for their novel approach named label smoothing regularization for outliers (LSRO). Their work shows that the usage of GAN-generated data can effectively improves the discriminative ability of the model. To address the problem of Scale-Adaptive Low Resolution Person Re-identification (SALR-REID), Wang et al.~\cite{DBLP:conf/ijcai/WangYYBS18} proposed a new framework named Cascaded Super-Resolution GAN (CSRGAN) which is composed of multiple SRGANs~\cite{Ledig_2017_CVPR} in series. Wei et al.~\cite{Wei_2018_CVPR} developed a new model called Person Transfer Generative Adversarial Network (PTGAN) to narrowed-down the domain gap that commonly exists between different datasets.

\subsection{Human Pose Estimation}
Human pose estimation is an significant problem that has enjoyed considerable attention in the are of computer vision and image processing. It has lots of applications such as video surveillance, criminal investigation via video searching and human-computer interaction.
Human pose estimation task aims to study the algorithms and systems that recover the pose of an articulated body, which consists of joints and rigid parts using image-based observations~\footnote{\url{https://en.wikipedia.org/wiki/Articulated_body_pose_estimation}}. The most common way to tackle this issue is to perform a single-person pose estimation for each detection~\cite{Cao_2017_CVPR}. For example, Johnson et al.~\cite{DBLP:conf/bmvc/JohnsonE10} developed an extension of the pictorial structure model~\cite{DBLP:journals/ijcv/FelzenszwalbH05} that incorporates richer models of appearance and prior over pose without introducing unacceptable computational expense. Tompson et al.~\cite{DBLP:conf/nips/TompsonJLB14} introudced a novel hybrid framework by combining a Convolutional Network Part-Detector with a MRF inspired  Spatial-Model to address the problem of articulated human pose estimation. which can utilize geometric relationships between body joint locations. Ouyang et al.~\cite{DBLP:conf/cvpr/OuyangCW14} proposed a novel model to extract the global and high-order human body articulation patterns from different aspects of information sources, such as mixture type, appearance score and deformation. Toshev and Szegedy~\cite{Toshev_2014_CVPR} formulated the pose estimation problem as a DNN-based regression problem towards body joints. Besides, a cascade of DNN-based pose predictors was proposed, which allows for increased precision of joint localization. Newell et al.~\cite{DBLP:conf/eccv/NewellYD16} introduced a new approach called "stacked hourglass" network to capture and consolidate information across all scales of the image. Enlightened by~\cite{DBLP:conf/eccv/NewellYD16}, a CNN based model~\cite{DBLP:conf/fgr/BelagiannisZ17} was proposed to regresse a heatmap representation for each body keypoint, which combines a feed forward module with a recurrent module. Based on CNN, Wei et al.~\cite{DBLP:conf/cvpr/WeiRKS16} designed a novel pose machine framework with a sequential architecture for learning image features and image-dependent spatial models.

In this paper, inspired by~\cite{Cao_2017_CVPR} that proposed an efficient approach to detect the 2D pose of multiple people in an image by using a non-parametric representation named PAFs, we utilize this technique to generate skeleton representations from a pose-rich dataset. These skeleton representations are treated as source of pose augmentation process. Besides, we introduce a novel GAN based model named siamese pose augmentation net to generate new images from skeleton samples motivated by the work~\cite{Yan:2017:SAM:3123266.3123277}, in which paired inputs containing human skeleton and appearance are used to new motion frames via conditional GAN (CGAN)~\cite{DBLP:journals/corr/MirzaO14}.

\section{Preliminaries}\label{sec:preliminaries}
In this section, the definitions of person re-identification and cross-view pose augmentation are given firstly, and then we review respectively the Generative Adversarial Networks (GAN) and Variational Autoencoder (VAE) in theory, which are two basic techniques of this work. Table~\ref{tab:notations} summarizes the mathematical notations used throughout this paper to facilitate the discussion of our work.

\begin{table}
    \caption{The summary of notations}
    \label{tab:notations}
    \setlength{\tabcolsep}{3pt}
    %\begin{tabular}{|p{50pt}|p{180pt}|}
    \begin{tabular}{|p{0.27\columnwidth}| p{0.62\columnwidth} |}
    \hline
    Notation&
    Definition\\
    \hline

    $\mathcal{V}$&
    a video surveillance system\\
    $V_i$&
    the $i$th camera view in $\mathcal{V}$\\
    $\mathcal{D}$&
    the image database of a video surveillance system\\
    $\mathcal{I}_i$&
    the image or video frames set of camera view $V_i$\\
    $\bm{I}^{(l)}_i$&
    the $l$-th sample in set $\mathcal{I}_i$ of camera view $V_i$\\
    $\Delta(\bm{I}^{(l)}_i)$&
    the person identification of the pedestrian in sample $\bm{I}^{(l)}_i$\\
    $\mathcal{R}$&
    the result set of person Re-Id task\\
    $\Omega$&
    a skeleton sample set\\
    $\bm{\omega}^{(\kappa)}$&
    the $\kappa$-th sample in $\Omega$\\
    $\mathfrak{M}_i$&
    the skeleton-to-appearance mapping for view $i$\\
    $\bm{\mathfrak{I}}^{(l)}_{i,k}$&
    a new visual sample generated by $l$-th image from $i$th camera view and $k$-th skeleton sample\\
    $\mathcal{A}_i$&
    the pose-augmented dataset of camera view $V_i$ \\
    $\bm{\alpha}^{(\kappa)}_i$&
    the $\kappa$-th samples in the pose-augmented dataset of camera view $V_i$\\
    $\bm{G}$&
    the generator of a $GAN$ \\
    $\bm{D}$&
    the discriminator of a $GAN$ \\
    $P_n(\bm{I})$&
    the natrual random distribution of $\bm{I}$ \\
    $\bm{\theta}_g$&
    the model parameters of a generator \\
    $\bm{\theta}_d$&
    the model parameters of a discriminator \\
    $\mathcal{L}_{GAN}$&
    the GAN loss \\
    $\tau$&
    the iterations number of training \\
    $\eta$&
    the learning rate \\
    $\bm{En}$&
    the encoder of a VAE \\
    $\bm{De}$&
    the decoder of a VAE \\
    $\bm{z}$&
    a latent variable \\
    $\bm{z}^{(\kappa)}_{\bm{\alpha}_1}$&
    the latent variable generated from $\bm{\alpha}_1$ \\
    $\bm{z}^{(\kappa)}_{\bm{\alpha}_2}$&
    the latent variable generated from $\bm{\alpha}_2$ \\
    $\bm{\theta}_{En}$&
    the model parameters of a encoder \\
    $\bm{\theta}_{De}$&
    the model parameters of a decoder \\
    $P(\bm{z})$&
    the distribution of $\bm{z}$ \\
    $P_{\bm{\theta}_{En}}(\bm{z}|\bm{I})$&
    the distribution estimated by the encoder \\
    $P_{\bm{\theta}_{De}}(\bm{\hat{I}}|\bm{z})$&
    the distribution estimated by the decoder \\
    $Div_{KL}$&
    Kullback-Leibler divergence\\
    $\mathcal{L}_{VAE}$&
    the VAE loss\\
    $N(\bm{0},\bm{I})$&
    the multivariate Gaussian distribution\\
    $\bm{\psi}$&
    the model parameters of alignment model\\
    $\bm{\ddot{I}}$&
    the visual representation of image $\bm{I}$ generated by CNN\\
    $\Join$&
    the vector concatenation operator\\
    $\bm{y}^{(\kappa)}_i$&
    the $\kappa$-th groundtruth sample of view $V_1$\\
    $\bm{\xi}^{(\kappa)}_i$&
    the representation that is concatenated by feature vector of skeleton sample $\bm{\omega}^{(\kappa)}_1$ and appearance sample $\bm{I}^{(\kappa)}_1$\\
    $\zeta$&
    the weight for L1 loss\\
    $\delta$&
    the threshold in the loss of Align model\\

    \hline
    \end{tabular}
    \label{tab1}
\end{table}

\subsection{Problem Definition}
Given a video frame dataset of a surveillance system and an image which contains a target pedestrian, a person re-identification (Re-Id) task aims to retrieve all videos or images captured by surveillance cameras, which contain the same person. In the real case, a surveillance system is equipped with a number of cameras in different positions. Videos recorded by these cameras have different height, views, light and etc.. If the query image and the target images are recorded by different cameras, the pedestrian recognition task can be called cross-view person Re-Id. To describe the cross-view person Re-Id task clearly, we give the formal definition as follows.

\begin{Definition}{\textbf{Person Re-identification.}}\label{def:person_reid}
    Suppose that there is a video surveillance system $\mathcal{V}$ equipped with $|\mathcal{V}|$ cameras, denoted by $\mathcal{V}=\{V_1,V_2,...,V_{|\mathcal{V}|}\}$. The set of video frames or images shot by camera $V_i$ is denoted as $\mathcal{I}_i=\{\bm{I}^{(1)}_i,\bm{I}^{(2)}_i,...,\bm{I}^{(n)}_i\}$. For the whole system, the database of surveillance images is denoted as $\mathcal{D}=\{\mathcal{I}_1,\mathcal{I}_2,...,\mathcal{I}_{|\mathcal{V}|}\}$. If a video frame or image $\bm{I}^{(l)}_i$ contains a pedestrian, the person identification is denoted as $\Delta(\bm{I}^{(l)}_i)$.

    Let $\mathcal{I}_i$ and $\mathcal{I}_j$ be the image sets of camera view $V_i$ and $V_j$ respectively. Given any one surveillance image $\bm{I}^{(\kappa)}_i \in \mathcal{I}_i$, the person re-identification problem is to search out the images of camera $V_j$ containing the same pedestrian, namely,
    \begin{equation}
        \mathcal{R}=\{\bm{I}^{(l)}_j|\Delta(\bm{I}^{(\kappa)}_i)=\Delta(\bm{I}^{(l)}_j), \bm{I}^{(\kappa)}_i \in \mathcal{I}_i, \bm{I}^{(l)}_j \in \mathcal{I}_j\}
    \end{equation}
    where $\mathcal{R}$ is the result set.
\end{Definition}

As a pedestrian in a real scenario always have different motions or poses (e.g., Bend, trot, waving hands, etc.) when he or she is walking on the road. In addition, the images of a person with the same posture look very different under disjoint cameras. That means for person Re-Id task, how to capture the common visual representations of a same pedestrian with many different motions or postures is a key challenge. However, the existing benchmarks such as VIPeR~\cite{Gray2007Evaluating}, CUHK03~\cite{Li_2014_CVPR}, Market-1501~\cite{DBLP:conf/iccv/ZhengSTWWT15} cannot provide abundant enough poses or motions of pedestrians to train the model. To overcome this challenge, in this work we introduce a cross-view pose augmentation approach for unsupervised cross-view person Re-Id by generating pose-rich visual samples to improve the performance of discrimination. The definition of cross-view pose augmentation is described in the following.

\begin{Definition}{\textbf{Cross-View Pose Augmentation.}}\label{def:cross_view_pose_aug}
    Cross-view pose augmentation is a specific data augmentation that aims to generate pose-rich samples from a skeleton sample set to augment the source cross-view dataset.

    Let $\mathcal{I}_i$ and $\mathcal{I}_j$ be two person image sets recorded by camera $V_i$ and $V_j$ respectively, and $\Omega = \{\bm{\omega}^{(1)},\bm{\omega}^{(2)},...,\bm{\omega}^{(|\Omega|)}\}$ be a skeleton sample set, $|\Omega|$ denotes the size of this set. Let $\mathfrak{M}_i$ and $\mathfrak{M}_j$ be a skeleton-to-appearance mapping from a pair of human appearance samples $\bm{I}^{(l)}_l$ and $\bm{I}^{(l)}_j$ and a skeleton sample $\bm{\omega}^{(\kappa)}$ to new produced paired samples $\bm{\mathfrak{I}}^{(l)}_{i,\kappa}$ and $\bm{\mathfrak{I}}^{(l)}_{j,\kappa}$ which have the same appearance of $\bm{I}^{(l)}_i$ and the same pose of $\bm{\omega}^{(\kappa)}$, namely,
    \begin{equation}
        \mathfrak{M}_i: (\bm{I}^{(l)}_i,\bm{\omega}^{(\kappa)}) \longrightarrow \bm{\mathfrak{I}}^{(l)}_{i,\kappa}
    \end{equation}
    \begin{equation}
        \mathfrak{M}_j: (\bm{I}^{(l)}_j,\bm{\omega}^{(\kappa)}) \longrightarrow \bm{\mathfrak{I}}^{(l)}_{j,\kappa}
    \end{equation}
    or they can be denoted as $\mathfrak{M}_i(\bm{I}^{(l)}_i,\bm{\omega}^{(\kappa)}) = \bm{\mathfrak{I}}^{(l)}_{i,\kappa}$ and $\mathfrak{M}_j(\bm{I}^{(l)}_j,\bm{\omega}^{(\kappa)}) = \bm{\mathfrak{I}}^{(l)}_{j,\kappa}$. Cross-view pose augmentation aims to generate new cross-view datasets $\mathcal{A}_i$ and $\mathcal{A}_j$ for view $V_i$ and $V_j$ from $\mathcal{I}_i$ and $\mathcal{I}_j$ by using skeleton-to-appearance mapping, formally,
    \begin{equation}
        \mathcal{A}_i = \{\mathcal{I}_i \cup \{\bm{\mathfrak{I}}^{(l)}_{i,\kappa}\}|\bm{\mathfrak{I}}^{(l)}_{i,\kappa}=\mathfrak{M}_i(\bm{I}^{(l)}_i,\bm{\omega}^{(\kappa)})\}
    \end{equation}
    \begin{equation}
        \mathcal{A}_j = \{\mathcal{I}_i \cup \{\bm{\mathfrak{I}}^{(l)}_{j,\kappa}\}|\bm{\mathfrak{I}}^{(l)}_{j,\kappa}=\mathfrak{M}_j(\bm{I}^{(l)}_j,\bm{\omega}^{(\kappa)})\}
    \end{equation}
    where $\{\bm{\mathfrak{I}}^{(l)}_{i,\kappa}\}$ and $\{\bm{\mathfrak{I}}^{(l)}_{j,\kappa}\}$ are the visual sample sets generated from skeleton samples and human appearance samples for view $V_i$ and $V_j$ respectively. It is obvious that the pose-augmented datasets $\mathcal{A}_i$ and $\mathcal{A}_j$ consist of the source dataset $\mathcal{I}_i$ and the new generated set $\{\bm{\mathfrak{I}}^{(l)}_{i,\kappa}\}$, $\mathcal{I}_j$ and the new generated set $\{\bm{\mathfrak{I}}^{(l)}_{j,\kappa}\}$ respectively. Therefore, datasets $\mathcal{A}_i$ and $\mathcal{A}_j$ cover more poses than $\mathcal{I}_i$ and $\mathcal{I}_j$, which are able to provide much more information of poses for pedestrian matching. To facilitate description, in this paper we denote the pose augmented dataset uniformly as $\mathcal{A}_i = \{\bm{\alpha}^{(1)}_i,\bm{\alpha}^{(2)}_i,...,\bm{\alpha}^{(|\mathcal{A}_i|)}_i\}$ and $\mathcal{A}_j = \{\bm{\alpha}^{(1)}_j,\bm{\alpha}^{(2)}_j,...,\bm{\alpha}^{(|\mathcal{A}_j|)}_j\}$ without any distinction of original images and newly synthesized samples.
\end{Definition}

To support the discussion, in the next two subsections we review the basic theories of generative adversarial networks (GAN) and variational autoencoder (VAE), which are unsed in the proposed approach.

\subsection{Review of Generative Adversarial Networks}
Generative Adversarial Networks (GAN for short)~\cite{DBLP:conf/nips/GoodfellowPMXWOCB14} is a generative model consisting of two components: a generator and a discriminator. It has superior performance in image generation, patterns of motion modeling, 3D objects reconstruction, etc. Specifically, the generator is to produce forged images according to real visual samples, and the discriminator's duty is to discriminate whether the inputs are forged by generator or from natural image distribution. The architecture of a GAN is illustrates in Fig.~\ref{fig:gan}.

\begin{figure}
    %\newskip\subfigtoppskip \subfigtopskip = -0.1cm
    \centering
    \includegraphics[width=1.0\linewidth]{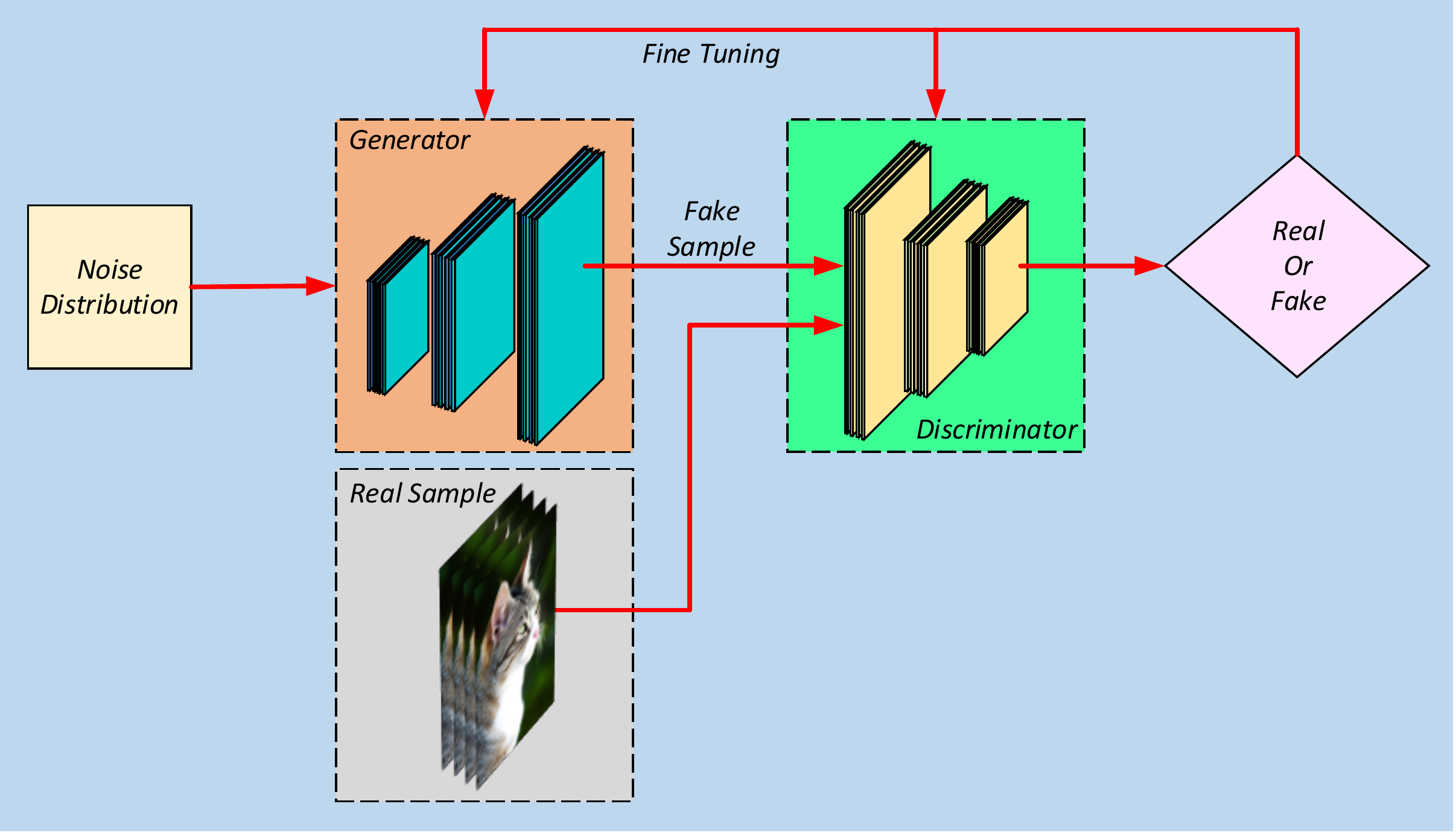}
    \vspace{-1mm}
    \caption{\small  The architecture of GAN. It has two neural networks: a generator and a discriminator. The former is to generate a image from the input, e.g., Gaussian noise signals. While the inputs of the latter are the real sample and the synthetic sample. The results of discrimination are used to fine tune these two components.}
    \label{fig:gan}
\end{figure}

For the convenience of discussion herein, the generator and discriminator are denoted as $\bm{G}$ and $\bm{D}$ respectively. In the whole training process, the generator $\bm{G}$ strenuously make the synthetic samples more similar to the real samples, while the discriminator $\bm{D}$ try its best to identify whether a input is from the generative model distribution or the natural distribution. In other words, this process is equivalent to a two-player zero-sum game. Along with the training the discriminator $\bm{D}$ and the generator $\bm{G}$ are diametrically against to each other. At last these two adversaries achieve a dynamic equilibrium: the generated sample is highly close to the natural distribution, while the discriminator $\bm{D}$ cannot distinguish true or false samples.

Let a real pedestrian sample $\bm{I}$ obeys natrual random distribution $P_n(\bm{I})$, and $\bm{z}$ be a randam sample from distribution $P_{\bm{z}}(\bm{z})$ in the form of a $\gamma$-dimension vector, namely $\bm{z}$ in $\mathbb{R}^\gamma$. The generator $\bm{G}(\bm{I};\bm{\theta}_{g})$ plays a role as a mapping from $\bm{z}$ to a synthetic sample $\bm{G}(\bm{z};\bm{\theta}_{g})$, and the generative distribution of $\bm{G}(\bm{z};\bm{\theta}_{g})$ is denoted as $P_{g}$. The discriminator $\bm{D}(\bm{I};\bm{\theta}_{d})$, on the other hand, receives the real sample $\bm{I}$ and the synthetic sample $\bm{G}(\bm{z};\bm{\theta}_{g})$ as input, and output the discriminant result $\bm{D}(\bm{G}(\bm{z};\bm{\theta}_{g});\bm{\theta}_{d})$ which is the probability that $\bm{G}(\bm{z};\bm{\theta}_{g})$ is synthesized from $\bm{G}$. This game process can be formulated as an minimax optimization of the following objective function $\mathcal{L}_{GAN}(\bm{G},\bm{D})$:

\begin{equation}\label{equ:loss_gan}
    \begin{split}
        \mathop{\arg}\mathop{\min}_{\bm{G}}\mathop{\max}_{\bm{D}}&\mathcal{L}_{GAN}(\bm{G},\bm{D})= \\ & \mathbb{E}_{\bm{I} \sim P_n(\bm{I})}[log\bm{D}(\bm{I};\bm{\theta}_{d})]+\\ & \mathbb{E}_{\bm{z} \sim P_{\bm{z}}(\bm{z})}[log(1-\bm{D}(\bm{G}(\bm{z};\bm{\theta}_{g});\bm{\theta}_{d}))]
    \end{split}
\end{equation}

where $\bm{\theta}_{g}$ and $\bm{\theta}_{d}$ are the network parameters of $\bm{G}$ and $\bm{D}$. $\mathbb{E}_{\bm{I} \sim P_n(\bm{I})}[\cdot]$ and $\mathbb{E}_{\bm{z} \sim P_{\bm{z}}(\bm{z})}[\cdot]$ are the expectation. Thus,

\begin{equation}\label{equ:expectation_I}
    \begin{split}
        \mathbb{E}_{\bm{I} \sim P_n(\bm{I})}&[log\bm{D}(\bm{I};\bm{\theta}_{d})]  = \\ & \int_{\bm{I}}{}P_n(\bm{I})log(\bm{D}(\bm{I};\bm{\theta}_{d}))\,d\bm{I}
    \end{split}
\end{equation}

\begin{equation}\label{equ:expectation_z}
    \begin{split}
        \mathbb{E}_{\bm{z} \sim P_{\bm{z}}(\bm{z})}&[log(1-\bm{D}(\bm{G}(\bm{z};\bm{\theta}_{g});\bm{\theta}_{d}))]  = \\ & \int_{\bm{z}}{}P_{\bm{z}}(\bm{z})log(1-\bm{D}(\bm{G}(\bm{z};\bm{\theta}_{g});\bm{\theta}_{d}))\,d\bm{z}
    \end{split}
\end{equation}

The generator $\bm{G}$ and discriminator $\bm{D}$ are trained in an alternate and iterative manner. For the generator $\bm{G}$, the training is to minimize the loss function to generate more authentic images to deceive the discriminator whose aim is to distinguish the synthetic samples. Therefore, for the discriminator $\bm{D}$, the objective is to maximize the loss. To make it more formal, the training of $\bm{G}$ and $\bm{D}$ can be denoted as:

\begin{equation}\label{equ:expectation_I}
    \begin{split}
        \mathop{\arg}\mathop{\min}_{\bm{G}}&\mathcal{L}_{GAN}(\bm{G},\bm{D}) = \\ & \int_{\bm{I}}{}P_n(\bm{I})log(\bm{D}(\bm{I};\bm{\theta}_{d}))\,d\bm{I}
    \end{split}
\end{equation}

\begin{equation}\label{equ:expectation_z}
    \begin{split}
        \mathop{\arg}\mathop{\max}_{\bm{D}}&\mathcal{L}_{GAN}(\bm{G},\bm{D}) = \\ & \int_{\bm{z}}{}P_{\bm{z}}(\bm{z})log(1-\bm{D}(\bm{G}(\bm{z};\bm{\theta}_{g});\bm{\theta}_{d}))\,d\bm{z}
    \end{split}
\end{equation}

During the course of the algorithm realized, the optimization of equation~\ref{equ:loss_gan} is implemented by using a stochastic gradient descent method. The gradient update steps are shown as follows:

for the discriminator $\bm{D}$:

\begin{equation}\label{equ:gradient_update_d}
    \begin{split}
        &\bm{\theta}^{\tau+1}_d = \bm{\theta}^{\tau}_d - \eta^\tau\nabla_{\bm{\theta}_d}\frac{1}{m}\\ &\sum_{\kappa=1}^m\left[log(\bm{D}(\bm{I}^{(\kappa)};\bm{\theta}^\tau_d))+log(1-\bm{D}(\bm{G}(\bm{z}^{(\kappa)};\bm{\theta}^\tau_g);\bm{\theta}^\tau_d))\right]
    \end{split}
\end{equation}

for the generator $\bm{G}$:

\begin{equation}\label{equ:gradient_update_d}
    \begin{split}
        \bm{\theta}^{\tau+1}_g =& \bm{\theta}^{\tau}_g - \eta^\tau\nabla_{\bm{\theta}_g}\frac{1}{m}\\ & \sum_{\kappa=1}^m\left[log(1-\bm{D}(\bm{G}(\bm{z}^{(\kappa)};\bm{\theta}^\tau_g);\bm{\theta}^{\tau+1}_d))\right]
    \end{split}
\end{equation}
where $m$ is the number of samples, $\eta$ is the learning rate, $\tau$ is the number of iterations. Obviously, in this process, the network parameters update via back-propagation only from discriminator, rather than modeling the reconstruction loss of the generator in a explicit manner.

In this work, GAN is considered as the main technique to construct the model of cross-view pose generation net (CPG-Net) that is to generate pose-rich images from existing datasets. Moreover, one of the key components in Cross-GAN model proposed by our previous work is based on GAN to learn a joint distribution of cross-view representations from multi-modal view-specific samples.

\subsection{Review of Variational Autoencoder}
Variational Autoencoder (VAE) is extended from autoencoder, which is introduced by Kingma et al.~\cite{DBLP:journals/corr/KingmaW13}. Like GAN, it is a commonly used deep generative model focused by lots of researchers in recent years.

A VAE generally consists of two parts: one is a encoder $\bm{En}(\bm{I})$, which maps a high-dimensional input $\bm{I}=(I^{(1)},I^{(2)},...,I^{(n)})$ to a low-dimensional latent variable $\bm{z}$, and the other is a decoder
$\bm{De}(\bm{z})$, which maps from a low-dimensional latent variable to a high-dimensional sample $\bm{\hat{I}}$ that is a reconstruction of $\bm{I}$, namely,

\begin{equation}\label{equ:vae_en_de}
    \begin{split}
        &\bm{z} \sim \bm{En}(\bm{I};\bm{\theta}_{En}) = P_{\bm{\theta}_{En}}(\bm{z}|\bm{I}),\\ & \bm{\hat{I}} \sim \bm{De}(\bm{z};\bm{\theta}_{De})=P_{\bm{\theta}_{De}}(\bm{\hat{I}}|\bm{z})
    \end{split}
\end{equation}
where $\bm{\theta}_{En}$ and $\bm{\theta}_{De}$ are the model parameters of encoder and decoder respectively. $P_{\bm{\theta}_{En}}(\bm{z}|\bm{I})$ is the posterior probability of $\bm{z}$ estimated by encoder, and $P_{\bm{\theta}_{De}}(\bm{\hat{I}}|\bm{z})$ is the posterior of $\bm{\hat{I}}$ by decoder.

\begin{figure}
    %\newskip\subfigtoppskip \subfigtopskip = -0.1cm
    \centering
    \includegraphics[width=1.0\linewidth]{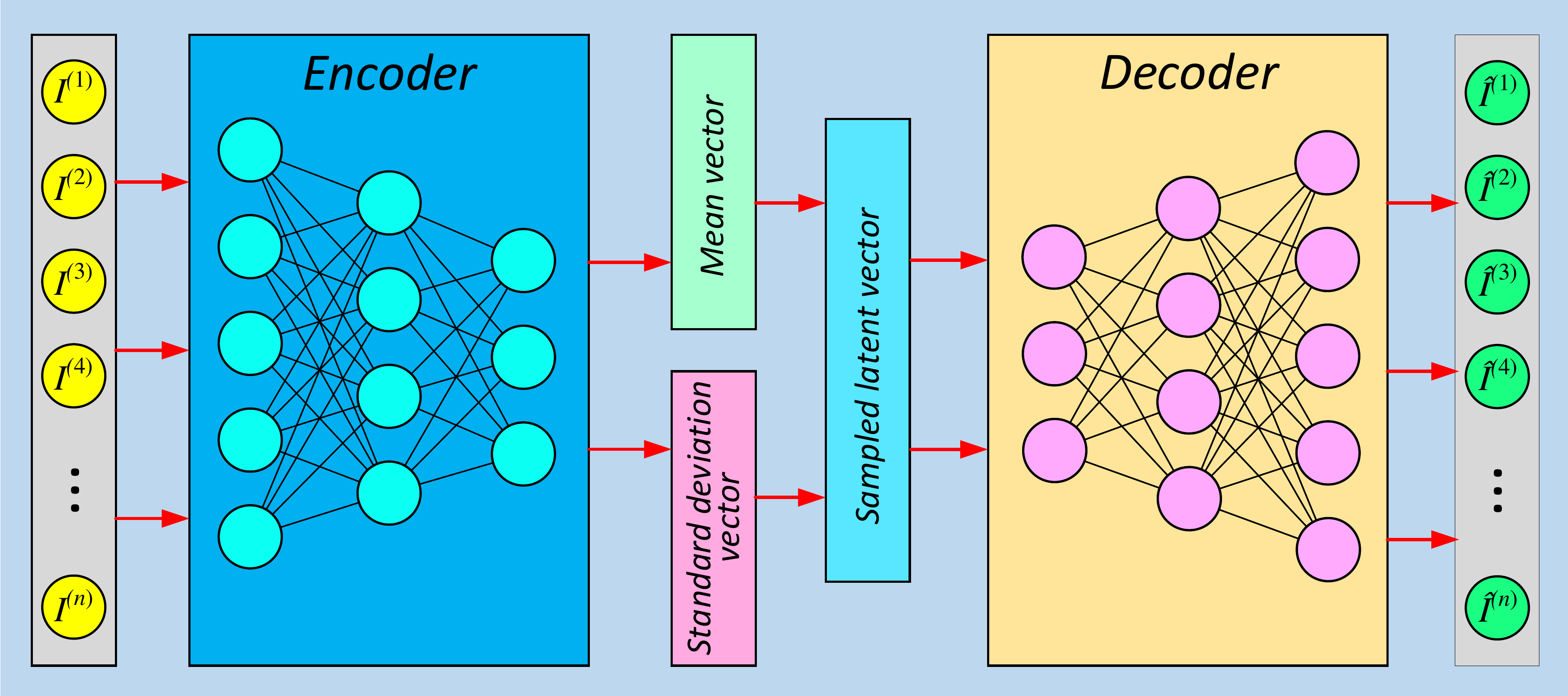}
    \vspace{-1mm}
    \caption{\small  The architecture of VAE. The encoder recieves an input $\bm{I}$ and maps it to a low-dimensional latent variable $\bm{z}$ by learning the posterior distribution of $\bm{z}$. The decoder is to generates samples $\bm{\hat{I}}$ as similar as possible to the original inputs from the latent variable.}
    \label{fig:gan}
\end{figure}

The optimization of VAE is to estimate a maximum likelihood probability distribution between the original input $\bm{I}$ and the reconstructed sample $\bm{\hat{I}}$ after mapping $\bm{I}$ to the latent variable, namely $P(\bm{I}) \simeq P_{\bm{\theta}_{De}}(\bm{z})$. The difference between these two distributions is measured by Kullback-Leibler (KL) divergence. Let $P(\bm{z})$ be the distribution of latent variable $\bm{z}$, $P(\bm{z})$ is assumed to be a Gaussian distribution with a mean of 0 and a variance of 1, i.e., $\bm{z} \sim N(0,1)$. Specifically, the KL divergence $Div_{KL}$ can be transformed according to Bayesian formula as follows:

\begin{equation}\label{equ:vae_kl}
    \begin{split}
        &Div_{KL}(P_{\bm{\theta}_{En}}(\bm{z}|\bm{I})||P(\bm{z})) \\ & = \int_{\bm{z}}{}P(\bm{z})log\left(\frac{P(\bm{z})}{P_{\bm{\theta}_{En}}(\bm{z}|\bm{I})}\right)\,d\bm{z} \\ & = \int_{\bm{z}}{}P(\bm{z})[log(P(\bm{z}))-log(P_{\bm{\theta}_{En}}(\bm{z}|\bm{I}))]\,d\bm{z} \\ & = \int_{\bm{z}}{}P(\bm{z})[log(P(\bm{z}))-log\left(\frac{P_{\bm{\theta}_{De}}(\bm{I}|\bm{z})P(\bm{I})}{P(\bm{z})}\right)]\,d\bm{z} \\ & = \int_{\bm{z}}{}P(\bm{z})[log(P(\bm{z}))-log(P_{\bm{\theta}_{De}}(\bm{I}|\bm{z}))- log(P(\bm{z}))]\,d\bm{z}\\ & \quad + log(P(\bm{I}))
    \end{split}
\end{equation}
this equation can be transformed by exchanging the left part and the right part as follows:

\begin{equation}\label{equ:vae_kl_ex}
    \begin{split}
        &log(\bm{I})-Div_{KL}(P(\bm{z}|\bm{I})||P_{\bm{\theta}_{En}}(\bm{z}|\bm{I})) \\ & = \int_{\bm{z}}{}P(\bm{z})log(P_{\bm{\theta}_{De}}(\bm{I}|\bm{z}))-Div_{KL}(P_{\bm{\theta}_{En}}(\bm{z}|\bm{I})||P(\bm{z}))
    \end{split}
\end{equation}

According to equation~\ref{equ:vae_kl_ex}, the loss function of VAE can be derived as follows:

\begin{equation}\label{equ:vae_loss}
    \begin{split}
        \mathcal{L}_{VAE}(\bm{En}(\bm{I};&\bm{\theta}_{En}),\bm{De}(\bm{z};\bm{\theta}_{De}))= \\ &+Div_{KL}(P_{\bm{\theta}_{En}}(\bm{z}|\bm{I})||P(\bm{z})) \\ &-\mathbb{E}_{\bm{z} \sim P_{\bm{\theta}_{En}}(\bm{z}|\bm{I})}[log(P_{\bm{\theta}_{De}}(\bm{I}|\bm{z}))]
    \end{split}
\end{equation}
where the first term on the right side of the above equation is KL divergence between the prior distribution of latent variable $P(\bm{z})$ and the encoded distribution $P_{\bm{\theta}_{En}}(\bm{z})$. The second term is the reconstruction loss which measures the difference between original inputs and reconstructed samples.

In our Cross-GAN model, a coupled VAE structure is applied to learning multi-modal distributions of cross-view visual samples without corresponding labeling. This component encodes the paired input into the
latent variables that are fed into the coupled GAN after a cross-view alignment process.

% \Figure[t!](topskip=0pt, botskip=0pt, midskip=0pt){fig1.png}
% {Magnetization as a function of applied field.
% It is good practice to explain the significance of the figure in the caption.\label{fig1}}

section{Methodology}
\label{sec:Methodology}
As pose is one of the significant modalities of the pedestrian images. which should be considered in the training process to capture the co-occurrence statistic patterns more precisely across different views. However, the existing benchmarks, like  VIPeR~\cite{Gray2007Evaluating}, CUHK03~\cite{Li_2014_CVPR}, Market-1501~\cite{DBLP:conf/iccv/ZhengSTWWT15} do not provide adequate enough poses variations in pedestrian images, which is a limitation for the model training. To overcome this challenge, we propose a novel \textbf{P}ose \textbf{A}ugmentation scheme for \textbf{C}ross-view person Re-Id based on \textbf{GAN} called \textbf{PAC-GAN}. To the best of our knowledge, we are the first to study the pose augmentation problem for unsupervised cross-view person Re-Id task. In this section, we introduce this framework in detail. At first, the overview of this framework is presented in subsection~\ref{subsec:overview}. Then we introudce the technique of skeleton generation applied in this work in subsection~\ref{subsec:skeleton-generation} and a novel CGAN based model for pose augmentation is proposed in subsection~\ref{subsec:CPG-Net}. In subsection~\ref{subsec:cross-gan}, we describe the unsupervised person Re-Id model that is integrated in this scheme.

\subsection{Overivew of our framework: PAC-GAN}
\label{subsec:overview}
\begin{figure*}
    %\newskip\subfigtoppskip \subfigtopskip = -0.1cm
    \centering
    \includegraphics[width=1.0\linewidth]{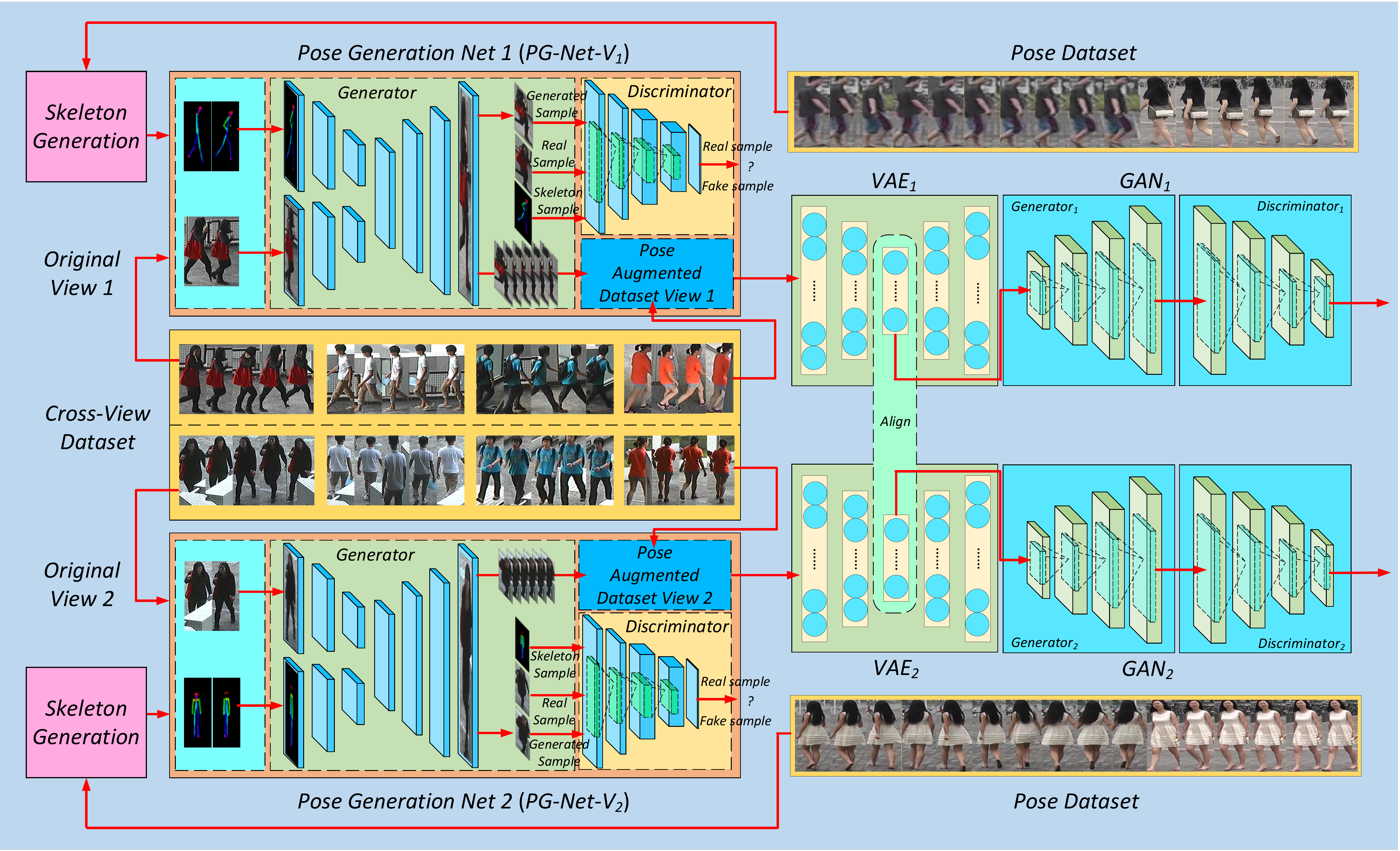}
    \vspace{-1mm}
    \caption{\small  The framework of PAC-GAN. This framework has two main models: (1) a coupled deep generative model named cross-vew pose generation net (CPG-Net) and (2) a coupled VAE and GAN model named Cross-GAN with cross-view alignment. The CPG-Net has two branches for two camera view $V_1$ and $V_2$, which are named Pose Generation Net for $V_1$ (PG-Net-$V_1$) and Pose Generation Net for $V_2$ (PG-Net-$V_2$). Weight-sharing strategy is applied across PG-Net-$V_1$ and PG-Net-$V_2$ to capture the co-occurrence visual patterns in cross-images to improve the performance of image generation. The input of them consists of two parts: one is the original pedestrian samples and the other is the skeleton sample produced by a skeleton generation process from a human motion image dataset. The new produced samples $\{\bm{\mathfrak{I}}_1\}$ and $\{\bm{\mathfrak{I}}_2\}$ have the poses from skeleton samples and the appearances from original pedestrian images, which are used to synthesize the pose augmented datasets $\mathcal{A}_1$ and $\mathcal{A}_2$ respectively. The unsupervised cross-view person Re-Id model consists of two branches of VAE and GAN denoted as (VAE$_1$, GAN$_1$) for view $V_1$ and (VAE$_2$, GAN$_2$) for view $V_2$. It receives the pose augmented datasets $\mathcal{A}_1$ and $\mathcal{A}_2$ as input to enhance the performance of identification. A cross-view alignment is employed on the latent variables of the coupled VAE to learn the statistic relationships between generative variables. The Crossing GAN is used to produce the joint view-invariant distribution of the inputs.}
    \label{fig:framework}
\end{figure*}

The framework of PAC-GAN consists of two main model: (1) a coupled deep generative model named \textbf{Cross-view} \textbf{P}ose \textbf{G}eneration \textbf{Net} (\textbf{CPG-Net}) which aims to generate new pose-rich pedestrian images from skeleton samples and person appearance samples, and (2) a coupled VAE and GAN model named \textbf{Cross-GAN}~\cite{DBLP:journals/ijon/ZhangWW19} to address cross-view person Re-Id problem in an unsupervised manner. The overview of PAC-GAN is illustrated in Fig.~\ref{fig:framework}.

\textbf{CPG-Net.} Specifically, the pose augmentation model shown in the left part of Fig.~\ref{fig:framework}, namely CPG-Net, is used to produce new samples with various poses according to inputs: the original cross-view samples and skeleton samples. This model is a coupled architecture, each of the nets corresponds to a camera view. For each view, a human motion or posture image dataset is employed as the source to generate skeleton samples. In our experiments, MARS dataset~\cite{DBLP:conf/eccv/ZhengBSWSWT16} is chosen to undertake this task. To achieve good performance, we utilize the image generation technique proposed by~\cite{Cao_2017_CVPR} to implement image-to-skeleton generation for each view. The key part of the pose generation net is a specific coupled GAN which is inspired by~\cite{Yan:2017:SAM:3123266.3123277}. Paired skeleton samples and original samples are fed into the generator with siamese structure, and the outputs are new images containing the poses from skeletons and the appearances from the original human images. The discriminator with a stack structure receives a triple: a real sample, a skeleton sample and a generated sample and discriminate that the input is real or not. The synthesized image set from PG-Net-$V_1$ and PG-Net-$V_2$ are denoted as $\{\bm{\mathfrak{I}}_1\}$ and $\{\bm{\mathfrak{I}}_2\}$ respectively. Each of them combines with its original datasets $\mathcal{I}_1$ and $\mathcal{I}_2$ as the final pose augmented datasets $\mathcal{A}_1$ and $\mathcal{A}_2$. It is crystal clear that comparing with the original cross-view dataset, $\mathcal{A}_1$ and $\mathcal{A}_2$ have much more various pose coverage.

\textbf{Cross-GAN.} The other main model of this framework, shown in the right part, is called Cross-GAN which is a unsupervised model to estimate a joint distribution of multi-modal cross-view visual samples to recognise the co-occurrence statistic patterns. Each of this coupled network consists of a paired VAE and GAN. The coupled VAE is used to encode the inputs into latent variables and a cross-view alignment is implemented over the latent variables to reduce the view disparity. This alignment operation generates a shared latent space by learnig the statistical correlation of the cross-view latent representation. The coupled GAN with weight-sharing is to capture the co-occurrence visual patterns appearing across the paired inputs.

In general, the combing CPG-Net with Cross-GAN can create a more powerful generative model for unsupervised cross-view person Re-Id. The CPG-Net is a competitive deep generative model to overcome the challenge of pose augmentation. It can strongly support Cross-GAN to learn the joint distribution of multi-modal cross-view images by producing numerous new pedestrian images with abundant enough pose coverage.

\subsection{Skeleton Generation Process}\label{subsec:skeleton-generation}
Before the pose-rich pedestrian images generation, the skeleton samples have to be produced. To tackle this issue efficiently, we adopt the pose estimation approach designed by~\cite{Cao_2017_CVPR}, in which a novel notion called part affinity fields is proposed. This technique uses a two branch CNN architecture to generate skeleton samples from images containing multiple person.

Specifically, let $\bm{\ddot{I}}$ be the visual representation of a image produced by the first 10 layers of VGG-19~\cite{DBLP:journals/corr/SimonyanZ14a}, one of the branch network denoted by $Conv_{\bm{\Lambda}}(\bm{\ddot{I}};\bm{\vartheta}^{(\tau)})$ learns a set of confidence maps $\bm{\Lambda}=\{\bm{\Lambda}_1,\bm{\Lambda}_2,...,\bm{\Lambda}_{|\bm{\Lambda}|}\}$ which represents body part locations. Each of the elements $\bm{\Lambda}_i \in \bm{\Lambda}$ corresponds one part of the body. And the other branch of this model, $Conv_{\bm{\Gamma}}(\bm{\ddot{I}};\bm{\varphi}^{(\tau)})$ is to output a set of vector fields $\bm{\Gamma}=\{\bm{\Gamma}_1,\bm{\Gamma}_2,...,\bm{\Gamma}_{|\bm{\Gamma}|}\}$ of part affinities to represent the association betwen parts, each $\bm{\Gamma}_j \in \bm{\Gamma}$ corresponds one limb of the body. $\bm{\vartheta}^{(\tau)}$ and $\bm{\varphi}^{(\tau)}$ are the network parameters of these two CNN in the stage $\tau$ respectively.

The skeleton generation process consists of several stages denoted as $ \tau \in \{1,2,...,n\}$ to refine the results in an iterative manner. In the first stage, namely $\tau=1$, the inputs of both the two branches are the original feature vectors $\bm{\ddot{I}}$, and the outputs are a set of confidence maps $\bm{\Lambda}^{(1)} = Conv_{\bm{\Lambda}}(\bm{\ddot{I}};\bm{\vartheta}^{(1)})$ and a set of affinity fields $\bm{\Gamma}^{(1)} = Conv_{\bm{\Gamma}}(\bm{\ddot{I}};\bm{\varphi}^{(1)})$. Then these two sets and the original visual representation are concatenated as a new vector denoted as $\bm{\ddot{I}} \Join \bm{\Lambda}^{(1)} \Join \bm{\Gamma}^{(1)}$ which is fed into the next stage, where $\Join$ is the vector concatenation operator. Overall, this iterative process can be formally described as follows:

%for confidence maps:
\begin{equation}\label{equ:img-to-sklt-iterations}
    \bm{\Lambda}^{(\tau)}=
    \begin{cases}
        Conv_{\bm{\Lambda}}(\bm{\ddot{I}};\bm{\vartheta}^{(\tau)}), & \tau = 1, \\
        Conv_{\bm{\Lambda}}(\bm{\ddot{I}} \Join \bm{\Lambda}^{(\tau-1)} \Join \bm{\Gamma}^{(\tau-1)};\bm{\vartheta}^{(\tau)}), & \tau \geq 2, \\
    \end{cases}
\end{equation}

%for affinity fields:
\begin{equation}\label{equ:img-to-sklt-iterations}
    \bm{\Gamma}^{(\tau)}=
    \begin{cases}
        Conv_{\bm{\Gamma}}(\bm{\ddot{I}};\bm{\varphi}^{(\tau)}), & \tau = 1, \\
        Conv_{\bm{\Gamma}}(\bm{\ddot{I}} \Join \bm{\Lambda}^{(\tau-1)} \Join \bm{\Gamma}^{(\tau-1)};\bm{\varphi}^{(\tau)}), & \tau \geq 2, \\
    \end{cases}
\end{equation}

Two loss functions are used to refine the confidence maps and affinity fields in the end of each stage. For each branch, a weighted $L_2$ loss between the outputs and the groundtruth is employed. In the stage $\tau$, the loss functions applied after two convolutional networks are shown respectively as follows:

\begin{equation}\label{equ:img-to-sklt-loss}
    \begin{split}
        \mathcal{L}_{\bm{\Lambda}}^{(\tau)}=\sum_{i=1}^{|\bm{\Lambda}|}\sum_{\bm{\pi}}^{}\bm{W}(\bm{\pi})* \parallel \bm{\Lambda}^{(\tau)}_i(\bm{\pi})-\bm{\widetilde{\Lambda}}^{(\tau)}_i(\bm{\pi})\parallel^2_2
    \end{split}
\end{equation}

\begin{equation}\label{equ:img-to-sklt-loss}
    \begin{split}
        \mathcal{L}_{\bm{\Gamma}}^{(\tau)}=\sum_{j=1}^{|\bm{\Gamma}|}\sum_{\bm{\pi}}^{}\bm{W}(\bm{\pi})* \parallel \bm{\Gamma}^{(\tau)}_j(\bm{\pi})-\bm{\widetilde{\Gamma}}^{(\tau)}_j(\bm{\pi})\parallel^2_2
    \end{split}
\end{equation}
where $\bm{\pi}$ is a location of an image and $\bm{W}(\bm{\pi})$ is the weight at the location $\bm{\pi}$, which is used to avoid penalizing the correct prediction during the training. $\bm{\widetilde{\Lambda}}^{(\tau)}_i$ and $\bm{\widetilde{\Gamma}}^{(\tau)}_j(\bm{\pi})$ are the groundtruth confidence map and affinity field at $\bm{\pi}$. Thus, for the overall process, the objective is the sum of loss in each branch:

\begin{equation}\label{equ:img-to-sklt-loss}
    \mathcal{L}_{overall}=\sum_{\tau=1}^{n}\left(\mathcal{L}_{\bm{\Lambda}}^{(\tau)}+\mathcal{L}_{\bm{\Gamma}}^{(\tau)}\right)
\end{equation}
where $n$ is the number of stages.

\subsection{The Pose Augmentation Model: CPG-Net}\label{subsec:CPG-Net}
We propose to synthesize new pose-rich samples by using a novel generative model named CPG-Net. The source of CPG-Net for each view consists of two parts: one is the samples from original dataset and the other is the skeleton samples generated from the skeleton generation process. The two branches of this coupled deep network are named \textbf{P}ose \textbf{G}eneration \textbf{Net} for view $V_1$ (\textbf{PG-Net-$\bm{V}_{\bm{1}}$}) and \textbf{P}ose \textbf{G}eneration \textbf{Net} for view $V_2$ (\textbf{PG-Net-$\bm{V}_{\bm{2}}$}) based on Conditional GAN (CGAN)~\cite{DBLP:journals/corr/MirzaO14}. The generator $\bm{G}$ outputs the synthesized samples which have the poses from the skeleton samples and the appearances from the original pedestrian images. The discriminator $\bm{D}$ is against to $\bm{G}$ by recognising the fake samples.

\begin{figure}
    %\newskip\subfigtoppskip \subfigtopskip = -0.1cm
    \centering
    \includegraphics[width=1.0\linewidth]{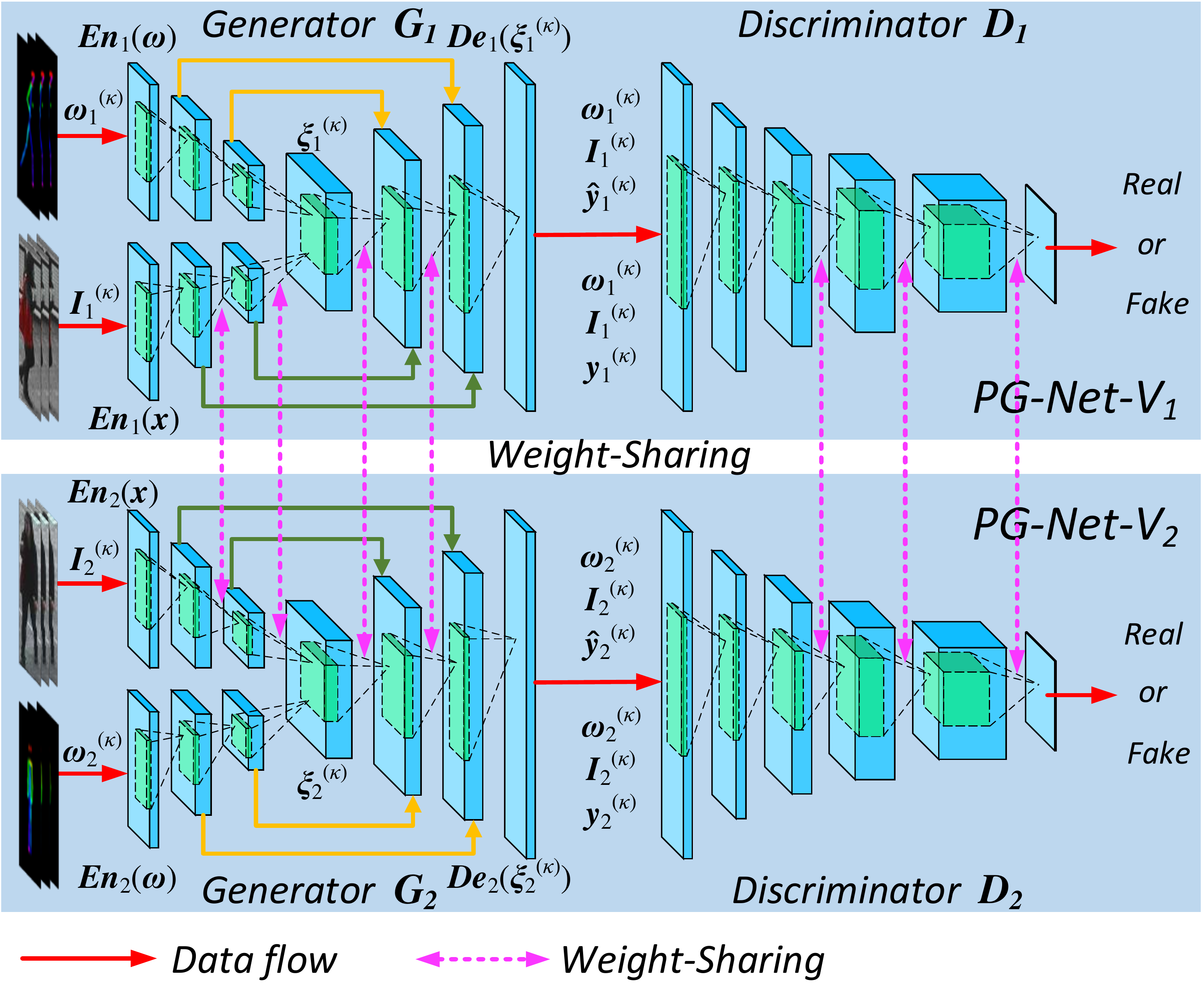}
    \vspace{-1mm}
    \caption{\small  The network architecture of the CPG-Net. It is a coupled network structure that is based on CGAN~\cite{DBLP:journals/corr/MirzaO14}:  the branch for $V_1$ is called PG-Net-$V_1$ and the other is named PG-Net-$V_2$. Each of the generators $\bm{G}_1$ and $\bm{G}_2$ has an encoder with siamese structure implemented by deep convolutional networks, whcih aim to generate representations of skeleton samples $\bm{\omega}^{(\kappa_1)}_1$ and $\bm{\omega}^{(\kappa_2)}_2$, as well as pedestrian appearance samples $\bm{I}^{(\kappa)}_1$ and $\bm{I}^{(\kappa)}_2$ respectively. For PG-Net-$V_1$ (the same as PG-Net-$V_2$), these two feature vectors are concatenated into a new code that is fed into the decoder which is implemented by deconvolution networks to generate the new samples $\bm{\mathfrak{I}}^{(\kappa)}_{1,\kappa_1}$ ($\bm{\mathfrak{I}}^{(\kappa)}_{2,\kappa_2}$ for PG-Net-$V_2$). The discriminator $\bm{D}_1$ receives two triples: the synthesized $\langle \bm{\omega}^{(\kappa)}_1, \bm{I}^{(\kappa)}_1, \bm{\mathfrak{I}}^{(\kappa)}_{1,\kappa_1} \rangle$ and the real $\langle \bm{\omega}^{(\kappa)}_1, \bm{I}^{(\kappa)}_1, \bm{\bm{y}}^{(\kappa)}_1 \rangle$, and then distinguishs the real from fake, where ${\bm{y}}^{(\kappa)}_1$ is the groundtruth sample. In order to generate the more authentic paired samples from two different camera views, we enforce last layers of $\bm{En}^{(\bm{I})}_1$ and $\bm{En}^{(\bm{I})}_2$, the first layers of $\bm{De}_1$ and $\bm{De}_1$, the last layers of discriminators $\bm{D}_1$ and $\bm{D}_2$ have duplicate network structure and parameters respectively, which is to capture the common high-level semantic features.}
    \label{fig:pg-net}
\end{figure}

\subsubsection{The Generator}

\textbf{Structure of Generator.} Inspired by~\cite{Yan:2017:SAM:3123266.3123277}, we design the generators of PG-Net-$V_1$ and PG-Net-$V_2$ denoted by $\bm{G}_1$ and $\bm{G}_2$ as a siamese structure "U-Net"~\cite{DBLP:conf/miccai/RonnebergerFB15} with weight-sharing across some layers, shown in Fig.~\ref{fig:pg-net}. Let $\Omega_1 = \{\bm{\omega}^{(1)}_1,\bm{\omega}^{(2)}_1,...,\bm{\omega}^{(|\Omega_1|)}_1\}$ and $\Omega_2 = \{\bm{\omega}^{(1)}_2,\bm{\omega}^{(2)}_2,...,\bm{\omega}^{(|\Omega_2|)}_2\}$ be the sets of skeleton samples for view $V_1$ and $V_2$ respectively, $\mathcal{I}_1=\{\bm{I}^{(1)}_1,\bm{I}^{(2)}_1,...,\bm{I}^{(|\mathcal{I}_1|)}_1\}$ and $\mathcal{I}_2=\{\bm{I}^{(1)}_2,\bm{I}^{(2)}_2,...,\bm{I}^{(|\mathcal{I}_2|)}_2\}$ be the sets of pedestrian appearance samples of view $V_1$ and $V_2$. For the generator $\bm{G}_1$ (same as $\bm{G}_2$), a siamese structure encoder namely $\bm{En}^{\bm{\omega}}_1$ and $\bm{En}^{\bm{I}}_1$ are utilized to model the skeleton samples and appearance samples. For a paired input $\langle\bm{\omega}^{(\kappa)}_1,\bm{I}^{(\kappa)}_1\rangle$, they are encoded by $\bm{En}^{\bm{\omega}}_1$ and $\bm{En}^{\bm{I}}_1$ in a convolution manner and then are concatenated into a new representation $\bm{\xi}^{(\kappa)}_1$, namely

\begin{equation}\label{equ:pg-net-encode}
    \bm{\xi}^{(\kappa)}_1 = \bm{En}^{\bm{\omega}}_1(\bm{\omega}^{(\kappa)}_1) \Join \bm{En}^{\bm{I}}_1(\bm{I}^{(\kappa)}_1)
\end{equation}
where $\Join$ is the concatenation operator. The decoder maps $\bm{\xi}^{(\kappa)}_1$ to a new sample $\bm{\mathfrak{I}}^{(\kappa)}_{i,\kappa_1} = \bm{De}_1(\bm{\xi}^{(\kappa)}_1)$ which has the same pose information of $\bm{\omega}^{(\kappa)}_1$ and inherits the appearance of $\bm{I}^{(\kappa)}_1$. Thus, the generators of $\bm{G}_1$ and $\bm{G}_2$ are represented formally as:

\begin{equation}\label{equ:pg-net-generator-1}
    \begin{split}
        \bm{G}_1(\bm{\omega}_1,\bm{\bm{I}}_1)=\bm{De}^{(n)}_1(\bm{De}^{(n-1)}_1(...\bm{De}^{(1)}_1(\bm{\xi}_1))), \\ \bm{\xi}_1 = \bm{En}^{\bm{\omega}(m)}_1(\bm{En}^{\bm{\omega}(m-1)}_1(...\bm{En}^{\bm{\omega}(1)}_2(\bm{\omega}_1))) \Join \\ \bm{En}^{\bm{I}(m)}_1(\bm{En}^{\bm{I}(m-1)}_1(...\bm{En}^{\bm{I}(1)}_1(\bm{I}_1)))
    \end{split}
\end{equation}

\begin{equation}\label{equ:pg-net-generator-2}
    \begin{split}
        \bm{G}_2(\bm{\omega}_2,\bm{\bm{I}}_2)=\bm{De}^{(n)}_2(\bm{De}^{(n-1)}_2(...\bm{De}^{(1)}_2(\bm{\xi}_2))), \\ \bm{\xi}_2 = \bm{En}^{\bm{\omega}(m)}_2(\bm{En}^{\bm{\omega}(m-1)}_2(...\bm{En}^{\bm{\omega}(1)}_2(\bm{\omega}_2))) \Join \\ \bm{En}^{\bm{I}(m)}_2(\bm{En}^{\bm{I}(m-1)}_2(...\bm{En}^{\bm{I}(1)}_2(\bm{I}_2)))
    \end{split}
\end{equation}
where $\bm{De}^{(i)}_1$, $\bm{De}^{(i)}_2$, $\bm{En}^{\bm{\omega}(i)}_1$, $\bm{En}^{\bm{I}(i)}_1$, $\bm{En}^{\bm{\omega}(i)}_2$ and $\bm{En}^{\bm{I}(i)}_2$ are the $i$-th layer of $\bm{De}_1$, $\bm{De}_2$, $\bm{En}^{\bm{\omega}}_1$, $\bm{En}^{\bm{I}}_1$, $\bm{En}^{\bm{\omega}}_2$ and $\bm{En}^{\bm{I}}_2$, respectively. $n$ and $m$ are respectively the number of their layers. All layers of the encoders are implemented by convolutional operation which encode the visual representations layer-by-layer from more detailed features to more abstract concept. The decoding process is the opposite, namely the decoder produce the new samples from $\bm{\xi}_1$ or $\bm{\xi}_2$ by a series of deconvolution operations. The first layer is to decode the high-level semantic concept and the last layer decode the low-level visual elements.

\textbf{Weight-Sharing in Generator.} As the cross-view images of a same person collected by disjoint cameras contains typically view-invariant high-level semantic features but different visual details, we use weight-sharing strategy in the encoder and decoder. Let $\bm{\theta}_{\bm{De}^{(i)}_1}$, $\bm{\theta}_{\bm{De}^{(i)}_2}$, $\bm{\theta}_{\bm{En}^{\bm{\omega}(i)}_1}$, $\bm{\theta}_{\bm{En}^{\bm{I}(i)}_1}$, $\bm{\theta}_{\bm{En}^{\bm{\omega}(i)}_2}$ and $\bm{\theta}_{\bm{En}^{\bm{I}(i)}_2}$ be the $i$-th layers parameters of the networks above, we propose to employ weight-sharing strategy across the last layers of $\bm{En}^{\bm{I}(i)}_1$ and $\bm{En}^{\bm{I}(i)}_2$, and the first layers of $\bm{De}_1$ and $\bm{De}_2$. In other words, the last layers of encoders $\bm{En}^{\bm{\omega}}_1$ and $\bm{En}^{\bm{\omega}}_1$, and the first layers of decoders $\bm{De}_1$ and $\bm{De}_2$ have duplicate network structure and parameters respectively, namely, $\bm{\theta}_{\bm{En}^{\bm{I}(i)}_1} = \bm{\theta}_{\bm{En}^{\bm{I}(i)}_2}, i \in \{1,2,...,p\}$, and $\bm{\theta}_{\bm{De}^{(j)}_1} = \bm{\theta}_{\bm{De}^{(j)}_2}, j \in \{1,2,...,q\}$, where $p$ and $q$ are the numbers of weight-shared layers in the encoders for appearance samples and the decoders, respectively.

\subsubsection{The Discriminator}
\textbf{Structure of Discriminator.} As shown in Fig.~\ref{fig:pg-net}, the discriminator in PG-Net-$V_1$ and PG-Net-$V_2$ recieves paired triples respectively: $\langle \bm{\omega}^{(\kappa)}_1, \bm{I}^{(\kappa)}_1, \bm{\mathfrak{I}}^{(\kappa)}_{1,\kappa_1} \rangle$ and $\langle \bm{\omega}^{(\kappa)}_1, \bm{I}^{(\kappa)}_1, \bm{y}^{(\kappa)}_1 \rangle$ for $V_1$, as well as $\langle \bm{\omega}^{(\kappa)}_2, \bm{I}^{(\kappa)}_2, \bm{\mathfrak{I}}^{(\kappa)}_{1,\kappa_1} \rangle$ and $\langle \bm{\omega}^{(\kappa)}_2, \bm{I}^{(\kappa)}_2, \bm{y}^{(\kappa)}_2 \rangle$ for $V_2$. Then it discriminates the input that is real or synthesized by implementing a stack structure. In other words, the triples are stacked as a 9-dimensional inputs and the discriminator extracts their features gradually and respectively, which are used to make a decision.

Let $\bm{D}^{(i)}_1$ and $\bm{D}^{(i)}_2$ be the $i$-th layer of the discriminator $\bm{D}_1$ and $\bm{D}_2$ for view 1 and view 2 respectively, $i \in \{1,2,...,r\}$, $r$ be the number of the layers. The $\bm{D}_1$ and $\bm{D}_2$ can be given by
\begin{equation}\label{equ:pg-net-D-1}
    \begin{split}
        \bm{D}_1(&\langle \bm{\omega}_1,\bm{I}_1,\bm{y}_1 \rangle) \\ & = \bm{D}^{(r)}_1(\bm{D}^{(r-1)}_1(...\bm{D}^{(1)}_1(\langle \bm{\omega}_1,\bm{I}_1,\bm{y}_1 \rangle)))
    \end{split}
\end{equation}

\begin{equation}\label{equ:pg-net-D-2}
    \begin{split}
        \bm{D}_2(&\langle \bm{\omega}_2,\bm{I}_2,\bm{y}_2 \rangle) \\ & = \bm{D}^{(r)}_2(\bm{D}^{(r-1)}_2(...\bm{D}^{(1)}_2(\langle \bm{\omega}_2,\bm{I}_2,\bm{y}_2 \rangle)))
    \end{split}
\end{equation}

\textbf{Weight-Sharing in Discriminator.} The cross-view discriminators are implemented by convolutional networks that extract the visual features layer-by-layer and at last output a probability score. The first layers of $\bm{D}_1$ and $\bm{D}_2$ extract low-level visual features and the last layers are used to perceive high-level semantic concept. Similar to the generator, we enforce the $\bm{D}_1$ and $\bm{D}_2$ have the identical network parameters to capture the co-occurrence abstract semantic concept from cross-image of the same pedestrian. On the other hand, weight-sharing can also contribute to reduce the number of model parameters. This constrain can be denoted as $\bm{\theta}_{\bm{D}^{(r-i)}_1} = \bm{\theta}_{\bm{D}^{(r-i)}_2}$, where $i = 1,2,...,s-1$, $s$ is the number of weight-sharing layers, $\bm{\theta}_{\bm{D}^{(r-i)}_1}$ and $\bm{\theta}_{\bm{D}^{(r-i)}_2}$ are the parameters of $(r-i)$-th layers of $\bm{D}_1$ and $\bm{D}_2$ respectively.

\subsubsection{The Loss of CPG-Net}
As discussed above, the proposed model CPG-Net is based on CGAN, in which each branch of network recieves paired inputs: skeleton sample $\bm{\omega}$ and pedestrian appearance samples pedestrian sample $\bm{I}$. According to the loss of CGAN, we give the loss functions of PG-Net-$V_1$ and PG-Net-$V_2$ as follows:
\begin{equation}\label{equ:pg-net-D-1}
    \begin{split}
        &\mathcal{L}_{PG\mbox{-}Net\mbox{-}V_1}(\bm{G}_1(\bm{\theta}_{G_1}),\bm{D}_1(\bm{\theta}_{\bm{D}_1})) \\ & = \mathbb{E}_{\langle \bm{\omega},\bm{I}\rangle \sim P(\langle\bm{\omega},\bm{I}\rangle),\bm{z}\sim P(\bm{z})}[log\bm{D}_1(\bm{\omega},\bm{I},\bm{y};\bm{\theta}_{\bm{D}_1})] \\ & + \mathbb{E}_{\langle \bm{\omega},\bm{I}\rangle \sim P(\langle\bm{\omega},\bm{I}\rangle),\bm{z}\sim P(\bm{z})}[log(1-\bm{D}_1(\bm{\omega},\bm{I},\bm{G}_1\\ &(\bm{\omega},\bm{I},\bm{z};\bm{\theta}_{G_1});\bm{\theta}_{\bm{D}_1}))]
    \end{split}
\end{equation}

\begin{equation}\label{equ:pg-net-D-2}
    \begin{split}
        &\mathcal{L}_{PG\mbox{-}Net\mbox{-}V_2}(\bm{G}_2(\bm{\theta}_{G_2}),\bm{D}_2(\bm{\theta}_{\bm{D}_2})) \\ & = \mathbb{E}_{\langle \bm{\omega},\bm{I}\rangle \sim P(\langle\bm{\omega},\bm{I}\rangle),\bm{z}\sim P(\bm{z})}[log\bm{D}_2(\bm{\omega},\bm{I},\bm{y};\bm{\theta}_{\bm{D}_2})] \\ & + \mathbb{E}_{\langle \bm{\omega},\bm{I}\rangle \sim P(\langle\bm{\omega},\bm{I}\rangle),\bm{z}\sim P(\bm{z})}[log(1-\bm{D}_2(\bm{\omega},\bm{I},\bm{G}_2\\ &(\bm{\omega},\bm{I},\bm{z};\bm{\theta}_{G_2});\bm{\theta}_{\bm{D}_2}))]
    \end{split}
\end{equation}
In order to generate better samples, we propose to combine the CGAN loss with L1 loss:
\begin{equation}\label{equ:pg-net-L1-1}
    \begin{split}
        &\mathcal{L}_{{L_1}_1}(\bm{G}_1(\bm{\theta}_{G_1})) \\ & = \mathbb{E}_{\langle \bm{\omega},\bm{I}\rangle \sim P(\langle\bm{\omega},\bm{I}\rangle),\bm{z}\sim P(\bm{z})}[\parallel \bm{y}-\bm{G}_1(\bm{\omega},\bm{I},\bm{z};\bm{\theta}_{G_1}) \parallel_1]
    \end{split}
\end{equation}
\begin{equation}\label{equ:pg-net-L1-2}
    \begin{split}
        &\mathcal{L}_{{L_1}_2}(\bm{G}_1(\bm{\theta}_{G_2})) \\ & = \mathbb{E}_{\langle \bm{\omega},\bm{I}\rangle \sim P(\langle\bm{\omega},\bm{I}\rangle),\bm{z}\sim P(\bm{z})}[\parallel \bm{y}-\bm{G}_2(\bm{\omega},\bm{I},\bm{z};\bm{\theta}_{G_2}) \parallel_1]
    \end{split}
\end{equation}
Therefore, the overrall loss function of the CPG-Net is:
\begin{equation}\label{equ:cpg-net-loss}
    \begin{split}
        \mathcal{L}_{CPG\mbox{-}Net}&(\bm{G}_1(\bm{\theta}_{G_1}),\bm{G}_2(\bm{\theta}_{G_2}),\bm{D}_1(\bm{\theta}_{\bm{D}_1}),\bm{D}_2(\bm{\theta}_{\bm{D}_1})) \\ & = \mathcal{L}_{PG\mbox{-}Net\mbox{-}1}(\bm{G}_1(\bm{\theta}_{G_1}),\bm{D}_1(\bm{\theta}_{\bm{D}_1})) \\ & + \mathcal{L}_{PG\mbox{-}Net\mbox{-}2}(\bm{G}_2(\bm{\theta}_{G_2}),\bm{D}_2(\bm{\theta}_{\bm{D}_2})) \\ & + \zeta(\mathcal{L}_{{L_1}_1}(\bm{G}_1(\bm{\theta}_{G_1})) + \mathcal{L}_{{L_1}_2}(\bm{G}_1(\bm{\theta}_{G_2})))
    \end{split}
\end{equation}
where $\zeta$ is the weight for L1 loss.

\subsection{Unsupervised Person Re-Id: Cross-GAN}\label{subsec:cross-gan}
The other part of the proposed scheme is called Cross-GAN that is introduced in our previous work~\cite{DBLP:journals/ijon/ZhangWW19}. It is a deep generative model to estimate the joint distributions of the multi-modal visual samples which have co-occurrence visual patterns for cross-view person Re-Id. This model is a coupled structure: each of the branches consists of a paired VAE and GAN, which are denoted as (VAE$_1$, GAN$_1$) corresponding to $V_1$ and (VAE$_2$, GAN$_2$) corresponding to $V_2$. The coupled VAEs aim to encode the cross-view input into a latent representation and a alignment process is appied over these latent variables of these two VAEs to reduce the view disparity. The coupled GANs with weight-sharing estimate the joint view-invariant distribution.

Let $\bm{\alpha}^{(\kappa)}_1 \in \mathcal{A}_1$ and $\bm{\alpha}^{(\kappa)}_2 \in \mathcal{A}_2$ are the paired samples of pose augmented datasets $\mathcal{A}_1$ and $\mathcal{A}_2$ which are generated by the pose augmentation model. To simplify the description, here we assume the number of samples in $\mathcal{A}_1$ and $\mathcal{A}_2$ are the same, namely $|\mathcal{A}_1|=|\mathcal{A}_2| = m$. The tuple $\langle\bm{\alpha}^{(\kappa)}_1,\bm{\alpha}^{(\kappa)}_2\rangle$ is fed into the coupled VAE and then encoded into latent variables $\bm{z}^{(\kappa)}_{\bm{\alpha}_1}$ and $\bm{z}^{(\kappa)}_{\bm{\alpha}_2}$. In our model, the encoders and decoders of the coupled VAE are implemented by multi-layered perceptions. The piror of the latent variables are assumed to be multivariate Gaussian, namely $P(\bm{z}_{\bm{\alpha}_1})=N_1(\bm{z}_{\bm{\alpha}_1},\bm{0},\bm{I})$ and $P(\bm{z}_{\bm{\alpha}_2})=N_2(\bm{z}_{\bm{\alpha}_2},\bm{0},\bm{I})$, and the variational approximate posterior be $N_1(\bm{z}_{\bm{\alpha}_1};\bm{\mu}^{(\kappa)}_1,\bm{\sigma}^{2(\kappa)}_1\bm{I})$ and $N_2(\bm{z}_{\bm{\alpha}_2};\bm{\mu}^{(\kappa)}_2,\bm{\sigma}^{2(\kappa)}_2\bm{I})$, where $N_1$ and $N_2$ are estimated by $V_1$ and $V_2$ encoders respectively, and $\bm{I}$ in here is identity matrix rather than a pedestrian image. Thus, according to the principle of VAE, the loss function for the $V_1$ is:

\begin{equation}\label{equ:vae_loss_view_1}
    \begin{split}
        &\mathcal{L}_{VAE}(\bm{\alpha}^{(\kappa)}_1) \\ & \simeq \frac{1}{2}\sum_{j=1}^{J}\left(1+log((\bm{\sigma}^{(\kappa)}_{1,j})^2)-((\bm{\mu}^{(\kappa)}_{1,j})^2)-((\bm{\sigma}^{(\kappa)}_{1,j})^2)\right), \\ & \bm{z}^{(\kappa)}_{\bm{\alpha}_2} = \bm{\mu}^{(\kappa)}_1+\bm{\sigma}^{(\kappa)}_1*\bm{\epsilon},\bm{\epsilon} \sim N(\bm{0},\bm{I})
    \end{split}
\end{equation}

and the loss function for the view 2 is:

\begin{equation}\label{equ:vae_loss_view_2}
    \begin{split}
        &\mathcal{L}_{VAE}(\bm{\alpha}^{(\kappa)}_2) \\ & \simeq \frac{1}{2}\sum_{j=1}^{J}\left(1+log((\bm{\sigma}^{(\kappa)}_{2,j})^2)-((\bm{\mu}^{(\kappa)}_{2,j})^2)-((\bm{\sigma}^{(\kappa)}_{2,j})^2)\right), \\ & \bm{z}^{(\kappa)}_{\bm{\alpha}_2} = \bm{\mu}^{(\kappa)}_2+\bm{\sigma}^{(\kappa)}_2*\bm{\epsilon},\bm{\epsilon} \sim N(\bm{0},\bm{I})
    \end{split}
\end{equation}

Therefore, the loss function of the coupled VAE with $m$ paired inputs $(\bm{\alpha}^{(\kappa)}_1,\bm{\alpha}^{(\kappa)}_2)$ is:
\begin{equation}\label{equ:vae_loss_view_2}
    \begin{split}
        \mathcal{L}_{VAE}&(\bm{\alpha}^{(\kappa)}_1,\bm{\alpha}^{(\kappa)}_2) \\ & = \frac{1}{m}\sum_{\kappa=1}^{m}\left(\mathcal{L}_{VAE}(\bm{\alpha}^{(\kappa)}_1)+\mathcal{L}_{VAE}(\bm{\alpha}^{(\kappa)}_2)\right)
    \end{split}
\end{equation}

To reduce the impact of view disparity, a cross-view alignment process is implemented over the latent variables of coupled VAE. This process can reveal underlying invariant properties among different views, and model the multi-modal distributions of cross-view data space. The loss of the alignment is shown as follows:
\begin{equation}\label{equ:align_loss}
    \mathcal{L}_{Align}(\bm{\psi})=\frac{1}{m}\sum_{\kappa=1}^{m}max\left(\parallel \bm{z}_{\bm{\alpha}_1}^{(\kappa)}-Align(\bm{z}_{\bm{\alpha}_2}^{(\kappa)};\bm{\psi}) \parallel^2_2,\delta \right)
\end{equation}
where $Align(\cdot)$ is the alignment model to produce a mapping across $P(\bm{z}_{\bm{\alpha}_1})$ and $P(\bm{z}_{\bm{\alpha}_2})$, $\bm{\psi}$ is the parameters of alignment model, and $\delta$ is the threshold.

The coupled GANs consisting of GAN$_1$ and GAN$_2$ are utilized to learn the joint distribution of cross-view images. The generators of two branches namely $\bm{G}_1(\bm{z}_{\bm{\alpha}_1};\bm{\theta}_{g_1})$ and $\bm{G}_2(\bm{z}_{\bm{\alpha}_2};\bm{\theta}_{g_2})$ decode the visual information from the latent representations $\bm{z}_{\bm{\alpha}_1}$ and $\bm{z}_{\bm{\alpha}_2}$ layer-by-layer via deconvolution, where $\bm{\theta}_{g_1}$ and $\bm{\theta}_{g_2}$ are the model parameters of $\bm{G}_1$ and $\bm{G}_2$. The discriminators $\bm{D}_1(\bm{G}_1(\bm{z}_{\bm{\alpha}_1};\bm{\theta}_{g_1});\bm{\theta}_{d_1})$ and $\bm{D}_2(\bm{G}_2(\bm{z}_{\bm{\alpha}_2};\bm{\theta}_{g_2});\bm{\theta}_{d_2})$ extract visual features from low-level to high-level by convolution, and learn the likelihood that the input is real or fake. $\bm{\theta}_{d_1}$ and $\bm{\theta}_{d_2}$ are the parameters of $\bm{D}_1$ and $\bm{D}_2$. Both of the generator and discriminator in this two branches have weight-sharing, which can reduce the model parameters and derive view-invariant features across $\bm{\alpha}_1$ and $\bm{\alpha}_2$. Therefore, the loss function of the coupled GAN is:
\begin{equation}\label{equ:gan-loss}
    \begin{split}
        \mathcal{L}_{GAN}&(
            \bm{G}_1(\bm{\theta}_{g_1}),
            \bm{G}_2(\bm{\theta}_{g_2}),
            \bm{D}_1(\bm{\theta}_{d_1}),
            \bm{D}_2(\bm{\theta}_{d_2})) \\ &
            = \frac{1}{m}\sum_{\kappa=1}^{m}\left[log\bm{D}_1(\bm{\alpha}^{(\kappa)}_1;\bm{\theta}_{d_1})\right] \\ &
            + \frac{1}{m}\sum_{\kappa=1}^{m}\left[log(1-\bm{D}_1(\bm{G}_1(\bm{z}^{(\kappa)}_{\bm{\alpha}_1};\bm{\theta}_{g_1});\bm{\theta}_{d_1})\right] \\ &
            + \frac{1}{m}\sum_{\kappa=1}^{m}\left[log\bm{D}_2(\bm{\alpha}^{(\kappa)}_2;\bm{\theta}_{d_2})\right] \\ &
            + \frac{1}{m}\sum_{\kappa=1}^{m}\left[log(1-\bm{D}_2(\bm{G}_2(\bm{z}^{(\kappa)}_{\bm{\alpha}_2};\bm{\theta}_{g_2});\bm{\theta}_{d_2})\right]
    \end{split}
\end{equation}
and the overall loss of Cross-GAN is:
\begin{equation}\label{equ:cross-gan-loss}
    \mathcal{L}_{Cross\mbox{-}GAN} = \mathcal{L}_{VAE}+\mathcal{L}_{Align}+\mathcal{L}_{GAN}
\end{equation}

\section{Experiments}\label{sec:experiments}
In this section, the datasets and evaluation protocol of experiments are introduced at first. Then we describe the implementation details of our method. We evaluate the effect of weight-sharing in CPG-Net on three benchmarks and discusse the comparison of the proposed approach and the semi/un-supervised and supervised state-of-the-arts. The results of experiments illustrate that our pose augmentation scheme for unsupervised person Re-Id can enhance the recognition accuracy effectively.

\subsection{Datasets and Evaluation Protocol}\label{subsec:datasets}
\begin{figure*}
    %\newskip\subfigtoppskip \subfigtopskip = -0.1cm
    \centering
    \includegraphics[width=1.0\linewidth]{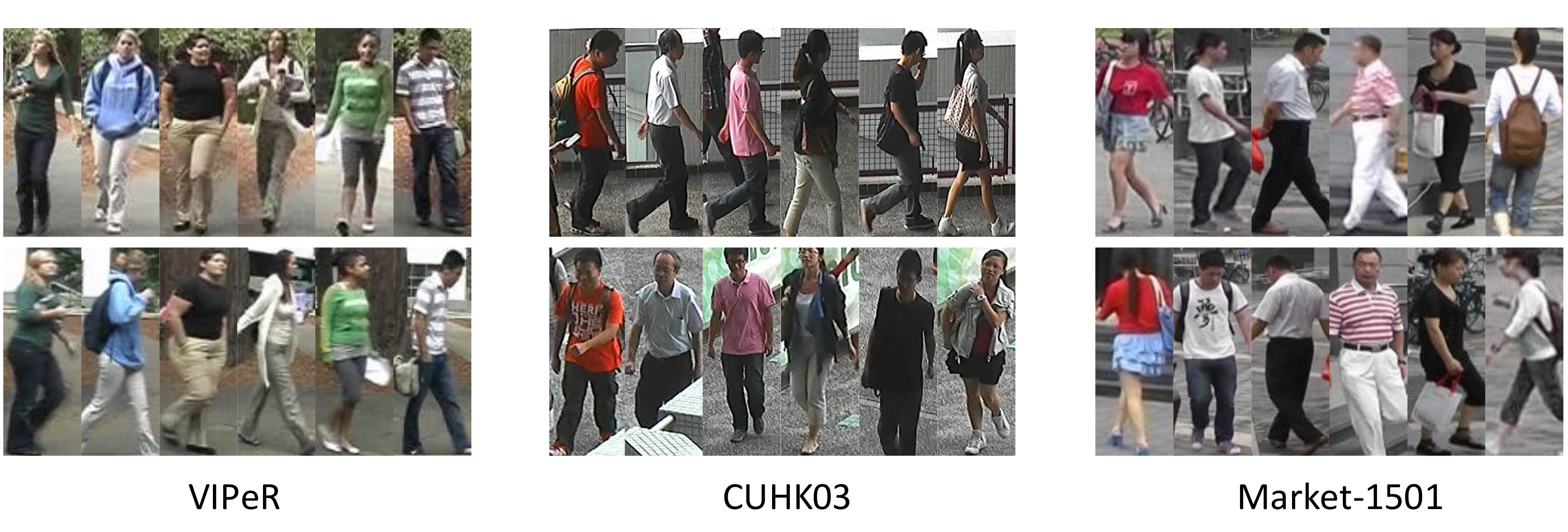}
    \vspace{-1mm}
    \caption{\small Some examples from person Re-Id datasets: VIPeR, CUHK03 and Market-1501}
    \label{fig:dataset}
\end{figure*}

\textbf{Datasets.} Our experiments are performed on three benchmarks: VIPeR~\cite{Gray2007Evaluating}, CUHK03~\cite{Li_2014_CVPR}, Market-1501~\cite{DBLP:conf/iccv/ZhengSTWWT15}. The detailed descriptions are given as follows:
\begin{itemize}
    \item \textbf{VIPeR} dataset consists of 1264 images of 632 pedestrians taken from arbitrary viewpoints under varying illumination conditions. Each pedestrian has two images. The size of each image is adjusted to 128$\times$48.
    \item \textbf{CUHK03} dataset contains 13164 cross-view samples of 1360 pedestrians collected by 6 surveillance cameras. The images of each pedestrian are taken from two different camera views. CUHK03 includes two subsets that contains manually labeled bounding boxes and automatically detected bounding boxes respectively. In this work, we perform experiments on labeled subset.
    \item \textbf{Market-1501} dataset includes 32643 images of 1501 pedestrians collected by 6 cameras. The boxes of pedestrians are obtained by a state-of-the-art detector of Deformable Part Model (DPM). The train set includes 750 identities and the testing set has 751 subjects.
\end{itemize}

Some examples of these three datasets are shown in Fig.~\ref{fig:dataset}. Each column indicates the images of the same pedestrian.

\textbf{Evaluation Protocol.} In our experiments, we use single-shot mode as the evaluation protocol. We calculate the rank that the query image is matched to the gallery images correctly. The rank-$k$ matching rate is the expectation of the matches at rank $k$, and the Cumulative Matching Characteristic (CMC) values at all ranks are reported.

\subsection{Implementation Details}\label{subsec:implement}
In this subsection, we introudce the implementation details of the main conmponents of this pose augmentation cross-view person Re-Id framework, namely skeleton generation model, CPG-Net and Cross-GAN respectively.

%\begin{table*}\scriptsize
\begin{table*}\tiny
    %\centering
    \caption{The Network Architecture of Coupled Generator of CPG-Net.}\label{tab:detail-coupled-generator}
    %\scalebox{0.9}
    \centering
    {
    \begin{tabular}{|l|c|c|c|}
    \hline
    \multirow{2}{*}{Layer} & \multicolumn{3}{c|}{\textbf{Encoder}}  \\
    \cline{2-4}
              & View 1 & View 2 &  Weight-sharing \\
    \hline
    1 & Conv (N=64, K=$5\times5$, S=2), LeakyReLU &  Conv (N=64, K=$5\times5$, S=2), LeakyReLU & No\\
     2 & Conv (N=128, K=$5\times5$, S=2), BN, LeakyReLU &  Conv (N=128, K=$5\times5$, S=2), BN, LeakyReLU & No\\
     3 & Conv (N=256, K=$5\times5$, S=2), BN, LeakyReLU &  Conv (N=256, K=$5\times5$, S=2), BN, LeakyReLU & No\\
     4 & Conv (N=512, K=$5\times5$, S=2), BN, LeakyReLU &  Conv (N=512, K=$5\times5$, S=2), BN, LeakyReLU & No\\
     5 & Conv (N=512, K=$5\times5$, S=2), BN, LeakyReLU &  Conv (N=512, K=$5\times5$, S=2), BN, LeakyReLU & Yes\\
     6 & Conv (N=512, K=$5\times5$, S=2), BN, LeakyReLU &  Conv (N=512, K=$5\times5$, S=2), BN, LeakyReLU & Yes\\
     7 & Conv (N=512, K=$5\times5$, S=2), BN, LeakyReLU &  Conv (N=512, K=$5\times5$, S=2), BN, LeakyReLU & Yes\\
     8 & Conv (N=512, K=$5\times5$, S=2), BN, LeakyReLU &  Conv (N=512, K=$5\times5$, S=2), BN, LeakyReLU & Yes\\
    \hline
    \multirow{2}{*}{Layer} & \multicolumn{3}{c|}{\textbf{Decoder}}  \\
    \cline{2-4}
     & View 1 & View 2 &  Weight-sharing \\
     \hline
   1 & FConv (N=512, K=$5\times5$, S=2), BN, Dropout (0.5), ReLU & FConv (N=512, K=$5\times5$, S=2), BN, Dropout (0.5), ReLU  &  Yes \\
   2 & FConv (N=1024, K=$5\times5$, S=2), BN, Dropout (0.5), ReLU & FConv (N=1024, K=$5\times5$, S=2), BN, Dropout (0.5), ReLU  &  Yes \\
   3 & FConv (N=1024, K=$5\times5$, S=2), BN, Dropout (0.5), ReLU & FConv (N=1024, K=$5\times5$, S=2), BN, Dropout (0.5), ReLU  &  Yes \\
   4 & FConv (N=1024, K=$5\times5$, S=2), BN, Dropout (0.5), ReLU & FConv (N=1024, K=$5\times5$, S=2), BN, Dropout (0.5), ReLU  &  Yes \\
   5 & FConv (N=1024, K=$5\times5$, S=2), BN, Dropout (0.5), ReLU & FConv (N=1024, K=$5\times5$, S=2), BN, Dropout (0.5), ReLU  &  No \\
   6 & FConv (N=512, K=$5\times5$, S=2), BN, Dropout (0.5), ReLU & FConv (N=512, K=$5\times5$, S=2), BN, Dropout (0.5), ReLU  &  No \\
   7 & FConv (N=256, K=$5\times5$, S=2), BN, Dropout (0.5), ReLU & FConv (N=256, K=$5\times5$, S=2), BN, Dropout (0.5), ReLU  &  No \\
   8 & FConv (N=128, K=$5\times5$, S=2), BN, Dropout (0.5), ReLU & FConv (N=128, K=$5\times5$, S=2), BN, Dropout (0.5), ReLU  &  No \\
    \hline
    \end{tabular}
    }
\end{table*}

\begin{table*}\scriptsize
    %\centering
    \caption{The Network Architecture of Coupled Discriminator of CPG-Net.}\label{tab:detail-coupled-discriminator}
    %\scalebox{0.9}
    \centering
    {
    \begin{tabular}{|l|c|c|c|}
    \hline
    \multirow{2}{*}{Layer} & \multicolumn{3}{c|}{\textbf{Discriminator}}  \\
    \cline{2-4}
              & View 1 & View 2 &  Weight-sharing \\
    \hline
    1 & Conv (N=64, K=$5\times5$, S=2), LeakyReLU &  Conv (N=64, K=$5\times5$, S=2), LeakyReLU & No\\
     2 & Conv (N=128, K=$5\times5$, S=2), BN, LeakyReLU &  Conv (N=128, K=$5\times5$, S=2), BN, LeakyReLU & No\\
     3 & Conv (N=256, K=$5\times5$, S=2), BN, LeakyReLU &  Conv (N=256, K=$5\times5$, S=2), BN, LeakyReLU & No\\
     4 & Conv (N=512, K=$5\times5$, S=2), BN, LeakyReLU &  Conv (N=512, K=$5\times5$, S=2), BN, LeakyReLU & No\\
     5 & Conv (N=512, K=$5\times5$, S=2), BN, LeakyReLU &  Conv (N=512, K=$5\times5$, S=2), BN, LeakyReLU & No\\
     6 & Conv (N=512, K=$5\times5$, S=2), BN, LeakyReLU &  Conv (N=512, K=$5\times5$, S=2), BN, LeakyReLU & Yes\\
    \hline
    \end{tabular}
    }
\end{table*}

%\begin{table*}\scriptsize
\begin{table*}\tiny
    %\centering
    \caption{The Network Architecture of Cross-GAN.}\label{tab:detail-cross-gan}
    %\scalebox{0.9}
    \centering
    {
    \begin{tabular}{|l|c|c|c|}
    \hline
    \multirow{2}{*}{Layer} & \multicolumn{3}{c|}{\textbf{Generator}}  \\
    \cline{2-4}
              & View 1 & View 2 &  Weight-sharing \\
    \hline
    1 & Conv (N=20, K=$5\times5$, S=1), BN, ReLU &  Conv (N=20, K=$5\times5$, S=1), BN, ReLU & Yes\\
     2 & Conv (N=20, K=$5\times5$, S=1), BN, ReLU &  Conv (N=20, K=$5\times5$, S=1), BN, ReLU & Yes\\
     3 & Conv (N=20, K=$5\times5$, S=1), BN, ReLU &  Conv (N=20, K=$5\times5$, S=1), BN, ReLU & Yes\\
     4 & Conv (N=20, K=$3\times3$, S=1), BN, ReLU &  Conv (N=20, K=$3\times3$, S=1), BN, ReLU & Yes\\
     5 & Conv (N=20, K=$3\times3$, S=1), BN &  Conv (N=20, K=$3\times3$, S=1), BN & No\\
    \hline
    \multirow{2}{*}{Layer} & \multicolumn{3}{c|}{\textbf{Discriminator}}  \\
    \cline{2-4}
     & View 1 & View 2 &  Weight-sharing \\
     \hline
   1 & Conv (N=20, K=$5\times5$, S=1), MAX-POOL (S=2), LeakyReLU & Conv (N=20, K=$5\times5$, S=1), MAX-POOL (S=2), LeakyReLU  &  No \\
   2 & Conv (N=20, K=$5\times5$, S=1), MAX-POOL (S=2), LeakyReLU & Conv (N=20, K=$5\times5$, S=1), MAX-POOL (S=2), LeakyReLU  &  No \\
   3 & Conv (N=20, K=$5\times5$, S=1), MAX-POOL (S=2), LeakyReLU & Conv (N=20, K=$5\times5$, S=1), MAX-POOL (S=2), LeakyReLU  &  No \\
   4 & FC (N=1024), ReLU & FC (N=1024), ReLU & No \\
   5 & FC (N=1024), Sigmoid & FC (N=1024), Sigmoid & Yes\\
    \hline
    \end{tabular}
    }
\end{table*}

\begin{figure}
    %\newskip\subfigtoppskip \subfigtopskip = -0.1cm
    \centering
    \includegraphics[width=1.0\linewidth]{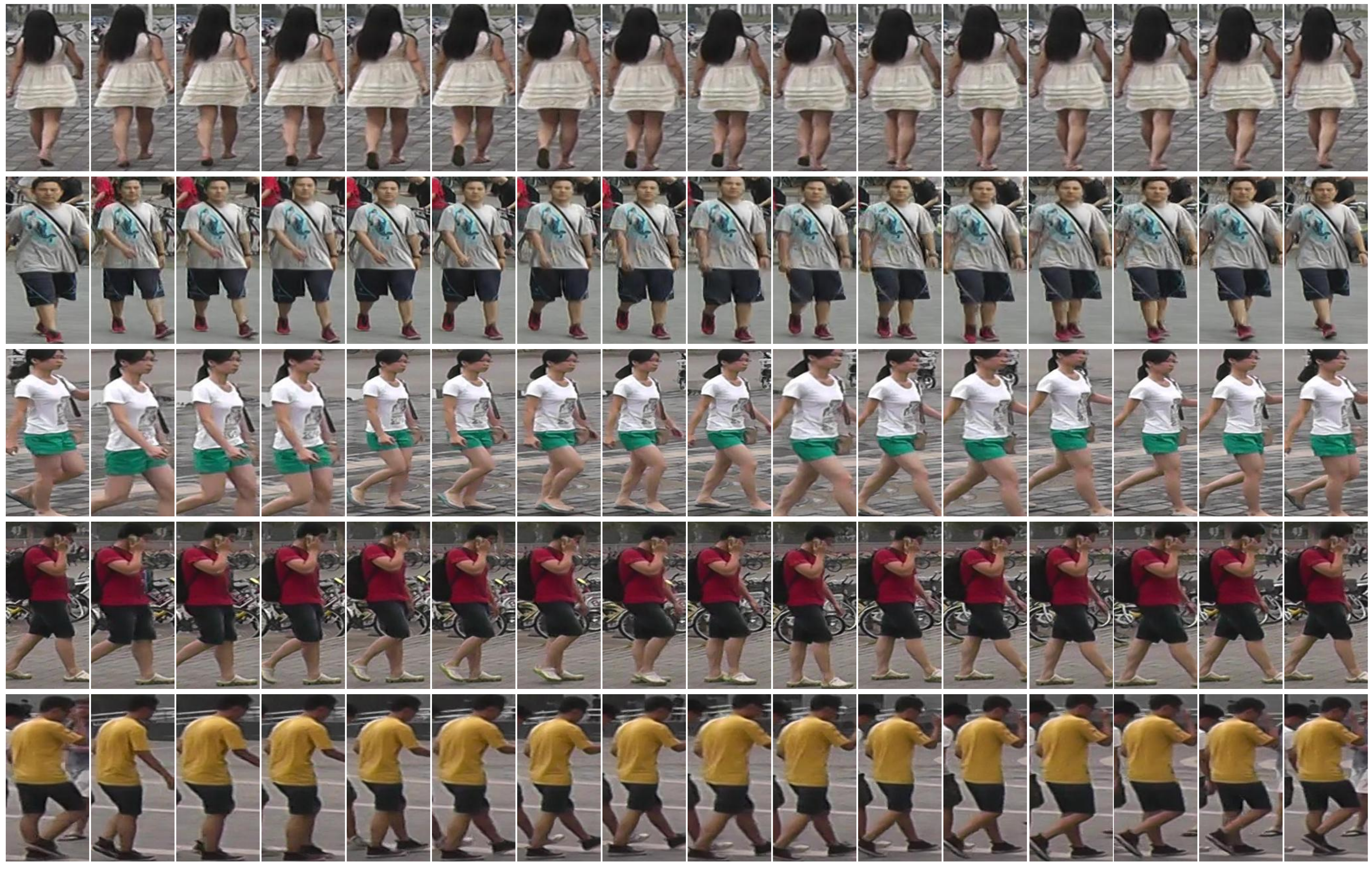}
    \vspace{-1mm}
    \caption{\small  The samples of MARS dataset. Mars has 1,261 identities and around 20,000 video sequences, which is one of the largest video person Re-Id dataset. Each row in this figure represents the images of a same person.}
    \label{fig:mars}
\end{figure}

\textbf{Skeleton Generation.} We apply the realtime human pose estimator method~\cite{Cao_2017_CVPR} to produce the skeleton samples from human images, which is the state-of-the-art of pose estimation. In our experiments, we use MARS dataset~\cite{DBLP:conf/eccv/ZhengBSWSWT16} as the data source to produce skeleton samples due to its wide rang of poses coverage. It is an extension of the Market-1501 dataset, which consists of 1,261 different pedestrians and around 20,000 video sequences. Some samples of MARS are shown in Fig.~\ref{fig:mars}. We feed the images of MARS into the pre-trained two-branches model mentioned above to produce the skeleton samples which is one of the sources of our pose augmentation model. But of course, MARS is not the only option.

\textbf{CPG-Net.} As discussed above, CPG-Net contains two network branches: PG-Net-$V_1$ and PG-Net-$V_2$, each of which is a CGAN based model. The encoder with siamese structure in coupled generator is implemented by the combination of Convolutional (Conv) layers, Batch Normalization (BN) layers and Leaky Rectified Unit (LeakyReLU) layers. The decoder of generator consists of Fractional Length Convolutional (FConv) layers, Batch Normalization (BN) layers, Dropout layers and ReLU layers. The details of network architecture are reported in Table~\ref{tab:detail-coupled-generator}.

The objective of coupled discriminators is to discriminate the natural triples and generated triples of two disjoint views. We implement this model by a stack structure, which includes Convolutional (Conv) layers, BN layers and LeakyReLU layers. The architecture description is shown in Table~\ref{tab:detail-coupled-discriminator}.

\textbf{Cross-GAN.} The generators in Cross-GAN have 5 convolutional layers and no spatial pooling layer is used in this model. This strategy allows the model to learn its own spatial down-sampling~\cite{DBLP:conf/eccv/ZhengBSWSWT16}. The discriminator is composed of Convolutional (Conv) layers, max pooling (MAX-POOL) layers, and LeakyReLU layers. In its last two layers, we use two fully connected (FC) layers with ReLU and Sigmoid respectively. The architecture of Cross-GAN is detailly reported in Table~\ref{tab:detail-cross-gan}.

\textbf{Weight-sharing.} By default, for CPG-Net, we set the number of weight-sharing layers in encoder and decoder is 4, which mainly aims to capture the common visual regions of cross-view images to improve the quality of synthesized pose-rich samples. In discriminator, the last two years share the weights. Likewise, the weight-sharing strategy is applied in Cross-GAN as well to enhance the performance of person matching. In the experiments discussed below, we evaluate how the weight-sharing strategy affect the performance of pose augmentation and pedestrians matching by varying the number of weight-sharing layers on different benchmarks.

\subsection{The Effect of Weight-Sharing in CPG-Net}\label{subsec:weight-sharing}

\begin{figure*}
    \newskip\subfigtoppskip \subfigtopskip = -0.1cm
    %\newskip\subfigcapskip \subfigcapskip = -0.2cm
    \begin{minipage}[b]{1\linewidth}
    \begin{center}
       %  \centering
         \subfigure[VIPeR]{
         \includegraphics[width=0.317\linewidth]{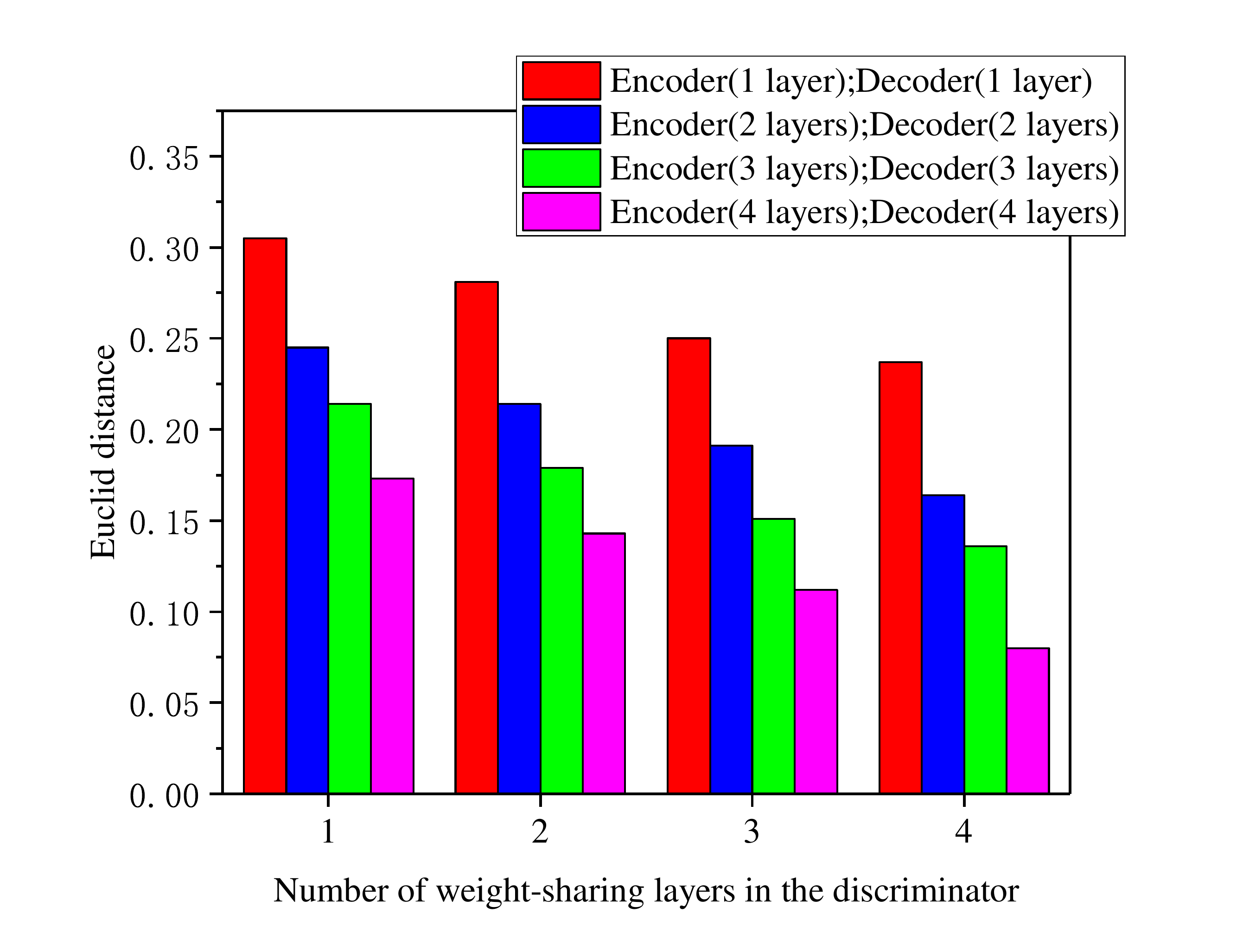}
         }
         \subfigure[CUHK03]{
         \includegraphics[width=0.317\linewidth]{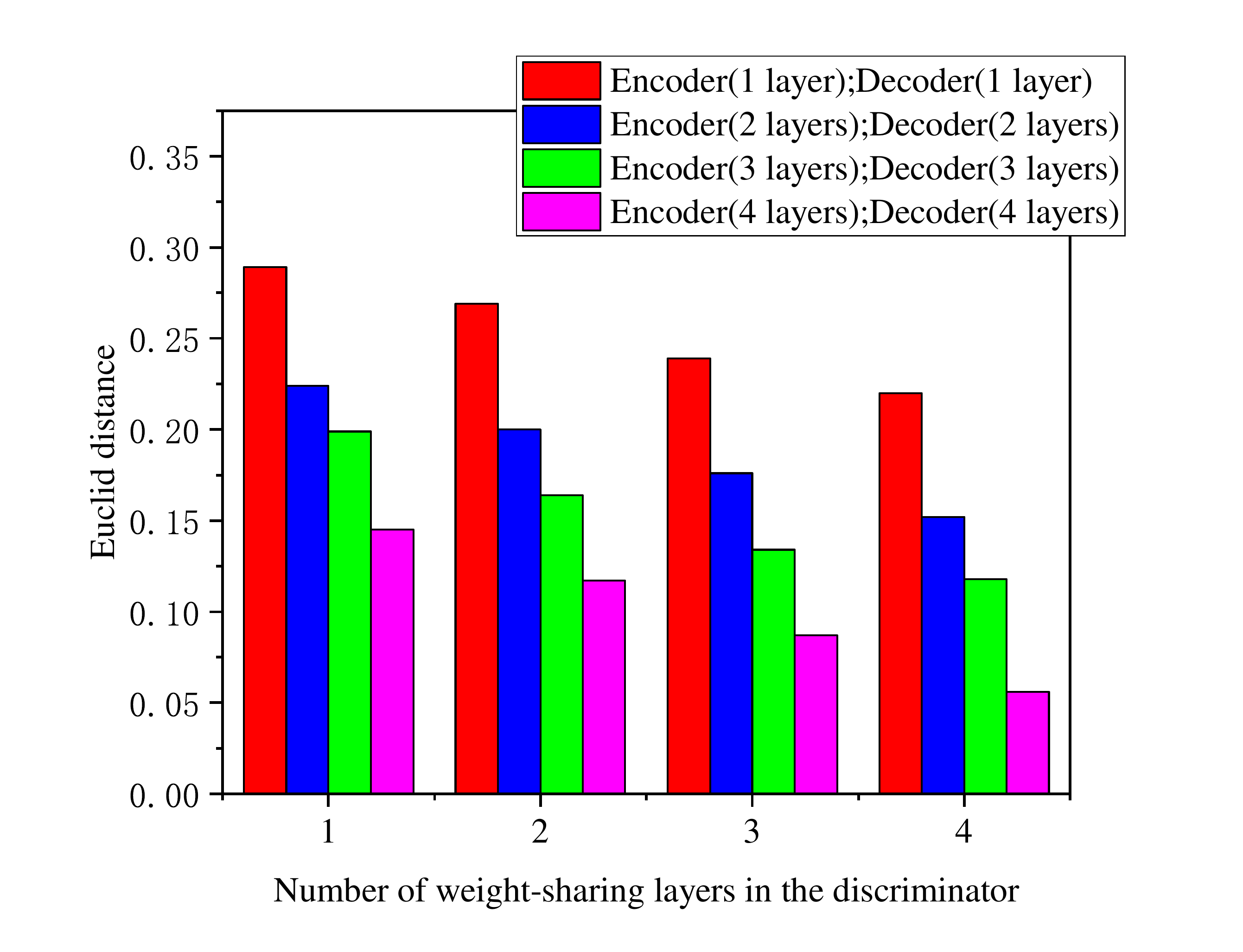}
         }
         \subfigure[Market-1501]{
         \includegraphics[width=0.317\linewidth]{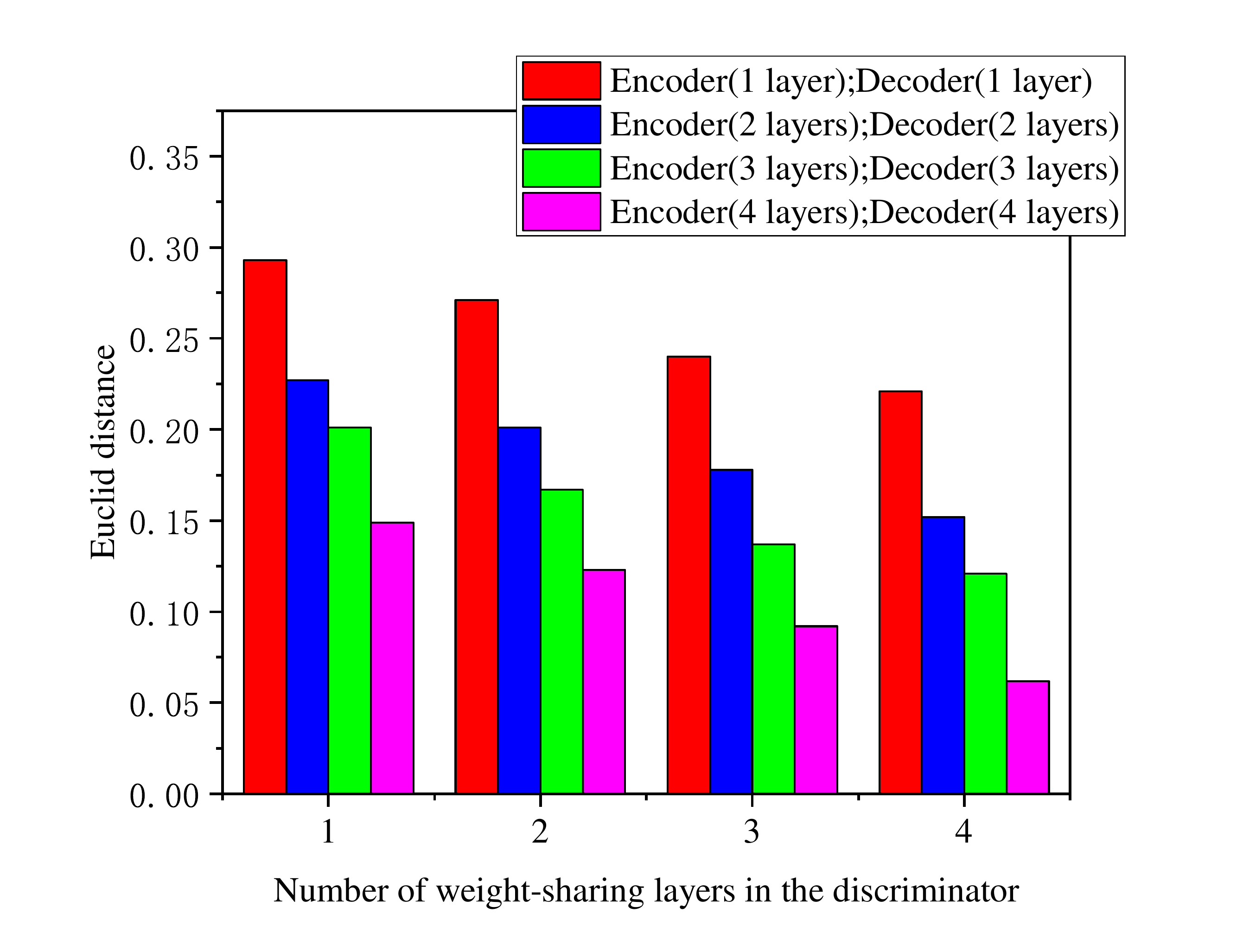}
         }
    \caption{The Euclidean distance measurements between two samples from VIPeR, CUHK03 and Market-1501 with different weight-sharing layers in CPG-Net. In this experiment, we set the numbers of weight-shaing layers in encoder and decoder are the same. The results show that the performance of image generation gradually increases with the rising the number of weight-sharing layers in generator (encoder and decoder). By comparion, the impact of weight-sharing in discriminator is relatively less.}
    \label{fig:weight-sharing}
    \end{center}
    \end{minipage}
\end{figure*}
We employ weight-sharing strategy on some specific layers of generator and discriminator in CPG-Net to learn co-occurrence visual pattern for pose-rich cross-view samples generation. The siamese encoders in the coupled generators with last layers weight-sharing encode the common visual regions of cross-view samples and the decoders with first layers weight-sharing synthesize pose-rich samples from the intermediate code. The coupled discriminators with last layer weight-sharing are able to capture the common visual patterns and prompt generators to produce more authentic images with new postures.

We evaluate the effect of weight-sharing in CPG-Net during the pose augmentation process by changing the number of weight-sharing layers of  generator and discriminator. The experiments are conducted on VIPeR, CUHK03 and Market-1501, and the results are shown in Fig.~\ref{fig:weight-sharing}.

It is obvious to find from Fig.~\ref{fig:weight-sharing} that the quality of generated cross-view samples is improved significantly with the increase of the number of weight-sharing layers in coupled generators (here we let the encoder and decoder have the same number of weight-sharing layers). By comparison, the performance of cross-view sample generation is not so much influenced by varying the number of weight-sharing layers in the coupled discriminators. For example, in Fig.~\ref{fig:weight-sharing} (a), when the encoder and decoder have one weight-sharing layer, the Euclidean distance between generated paired samples $\bm{I}_1$ and $\bm{I}_2$ decreases from 0.305 to 0.237 with the number of weight-sharing in discriminator from 1 to 4 (illustrated by the red bar in each column), which is evidently less than the variation (from 0.305 to 0.173) with the change of number of weight-sharing layers in generator (shown by the red bar and magenta bar in column 1). The results on the other two datasets also have the same trend, shown in Fig.~\ref{fig:weight-sharing} (b) and Fig.~\ref{fig:weight-sharing} (c) respectively. Although increasing the number of weight-sharing layers of the coupled discriminator contributes little for performance improvement of cross-view sample generation, this strategy still can reduce the number of network parameters in discriminative model to improve the training efficiency.

\subsection{Comparison with State-of-the-arts}\label{subsec:comparison}
In the following we compare the proposed approach PAC-GAN and Cross-GAN~\cite{DBLP:conf/eccv/ZhengBSWSWT16} with the following state-of-the-arts semi/un-supervised and supervised approaches on VIPeR, CUHK03 and Market-1501. The semi/un-supervised approaches include: SDALF~\cite{DBLP:conf/cvpr/FarenzenaBPMC10}, eSDC~\cite{Zhao_2013_CVPR}, t-LRDC~\cite{DBLP:journals/pami/ZhengGX16}, OSML~\cite{Bak_2017_CVPR}, LSRO~\cite{Zheng_2017_ICCV}, CAMEL~\cite{Yu_2017_ICCV}, UMDL~\cite{Peng_2016_CVPR}, BoW~\cite{DBLP:conf/iccv/ZhengSTWWT15} and PUL~\cite{DBLP:journals/tomccap/FanZYY18}. The supervised approaches include: DM$^3$~\cite{DBLP:journals/tcyb/0007H00JLS18}, DeepList~\cite{DBLP:journals/tcsv/WangWGSH17}, DDDM~\cite{DBLP:journals/tmm/WangHLYJYCL16}, Locally-Aligned~\cite{Li_2013_CVPR_Local}, JointRe-id~\cite{Ahmed_2015_CVPR}, SCSP~\cite{Chen2016Similarity}, Multi-channel~\cite{Cheng_2016_CVPR}, DNSL~\cite{Zhang_2016_CVPR}, JSTL~\cite{Xiao_2016_CVPR}, SI-CI~\cite{Wang_2016_CVPR}, S-CNN~\cite{Varior2016Gated}, SpindleNet~\cite{Zhao_2017_CVPR}, Part-Aligned~\cite{Zhao_2017_ICCV}, S-LSTM~\cite{DBLP:conf/eccv/VariorSLXW16}, E-Metric~\cite{DBLP:conf/eccv/ShiYZLLZL16}, Deep-Embed~\cite{Lin2017Deep}, SSM~\cite{Bai_2017_CVPR}, MSCAN~\cite{Li_2017_CVPR}, CADL~\cite{Lin_2017_CVPR}, LADF~\cite{Li_2013_CVPR}, XQDA~\cite{Liao2015Person}, OL-MANS~\cite{Zhou2017Efficient}, SalMatch~\cite{Zhao_2013_ICCV} and PDC~\cite{Su_2017_ICCV}. Note that not all the approaches above report the results of experiments in all three datasets.

\begin{figure*}
    \newskip\subfigtoppskip \subfigtopskip = -0.1cm
    %\newskip\subfigcapskip \subfigcapskip = -0.2cm
    \begin{minipage}[b]{1\linewidth}
    \begin{center}
       %  \centering
         \subfigure[VIPeR]{
         \includegraphics[width=0.317\linewidth]{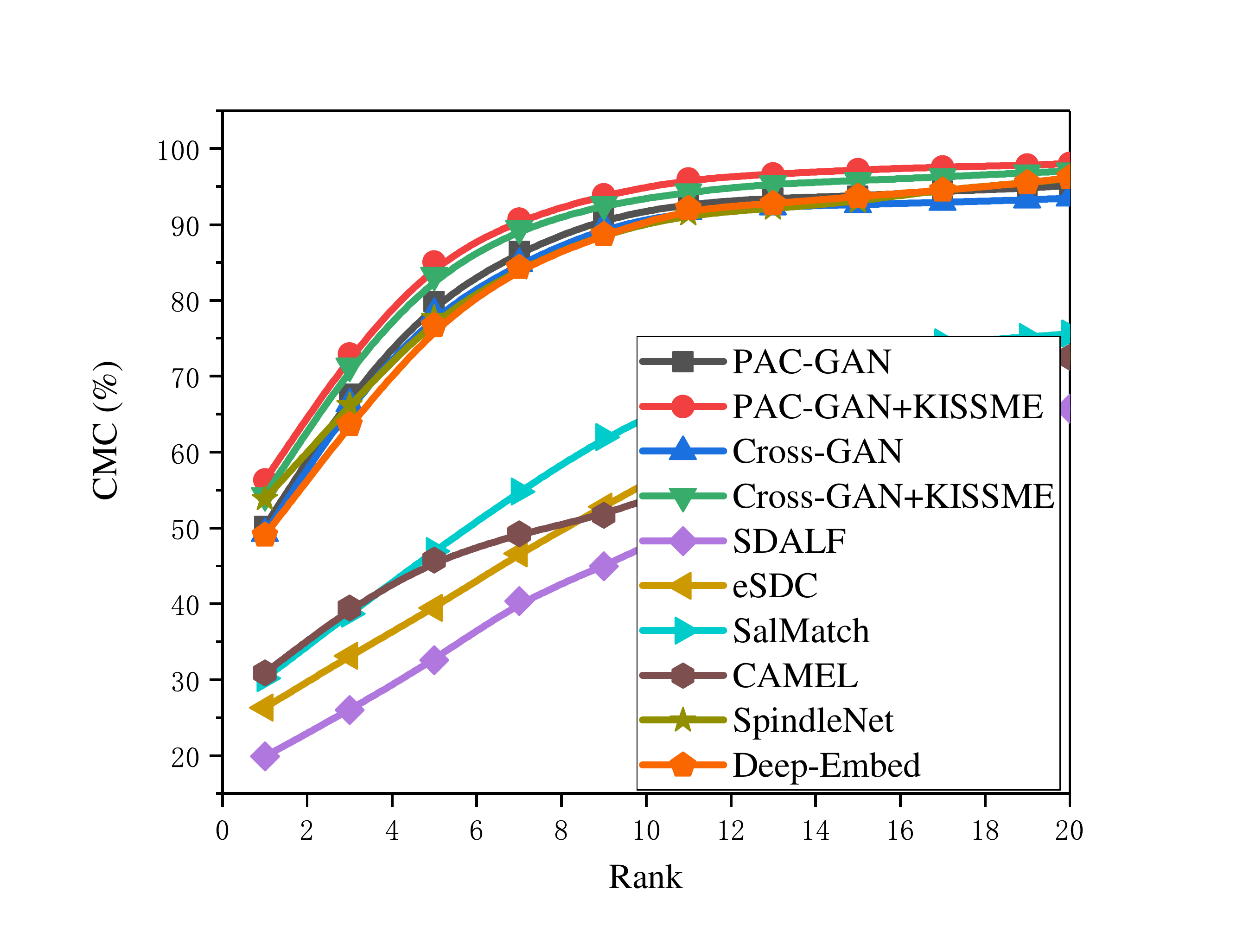}
         }
         \subfigure[CUHK03]{
         \includegraphics[width=0.317\linewidth]{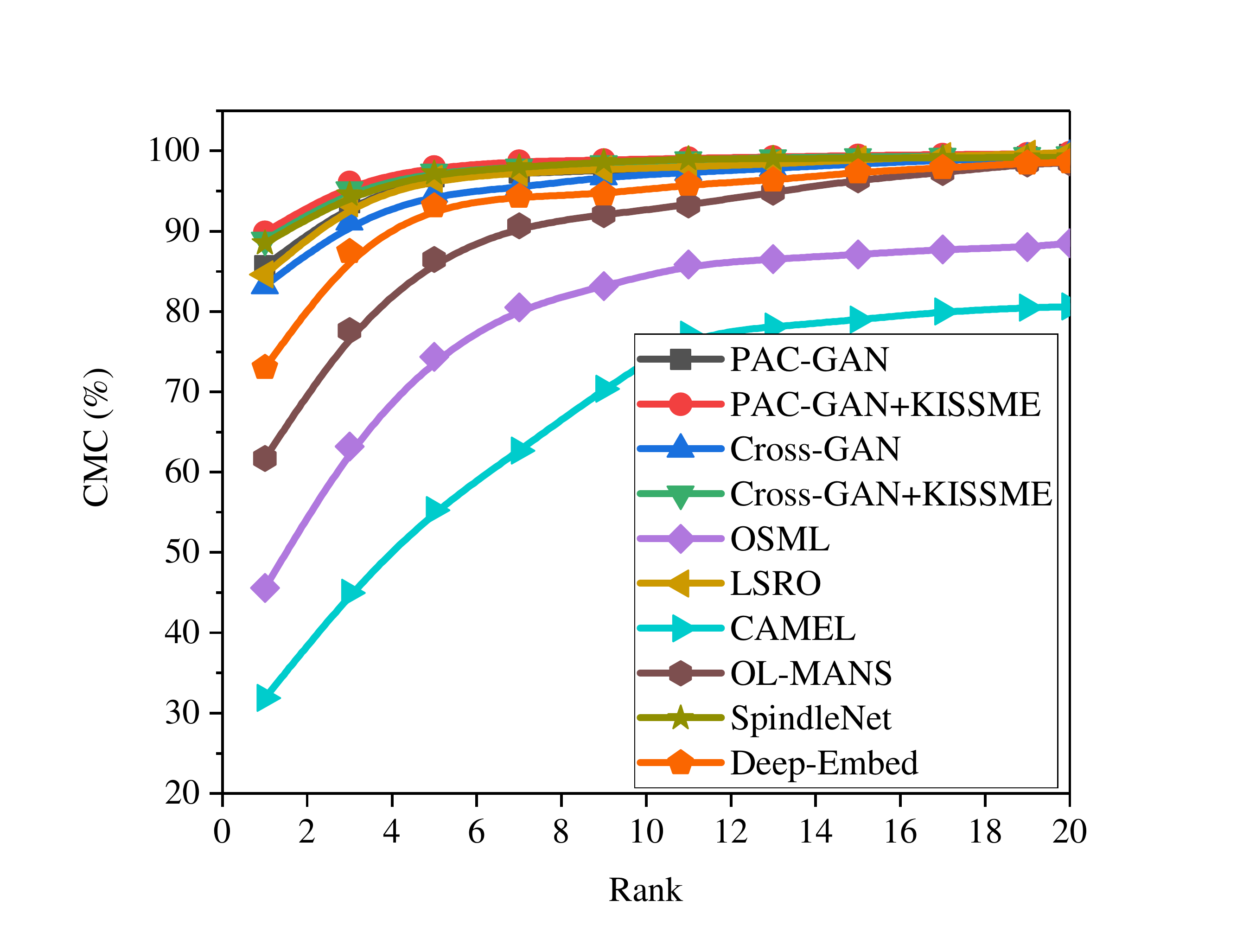}
         }
         \subfigure[Market-1501]{
         \includegraphics[width=0.317\linewidth]{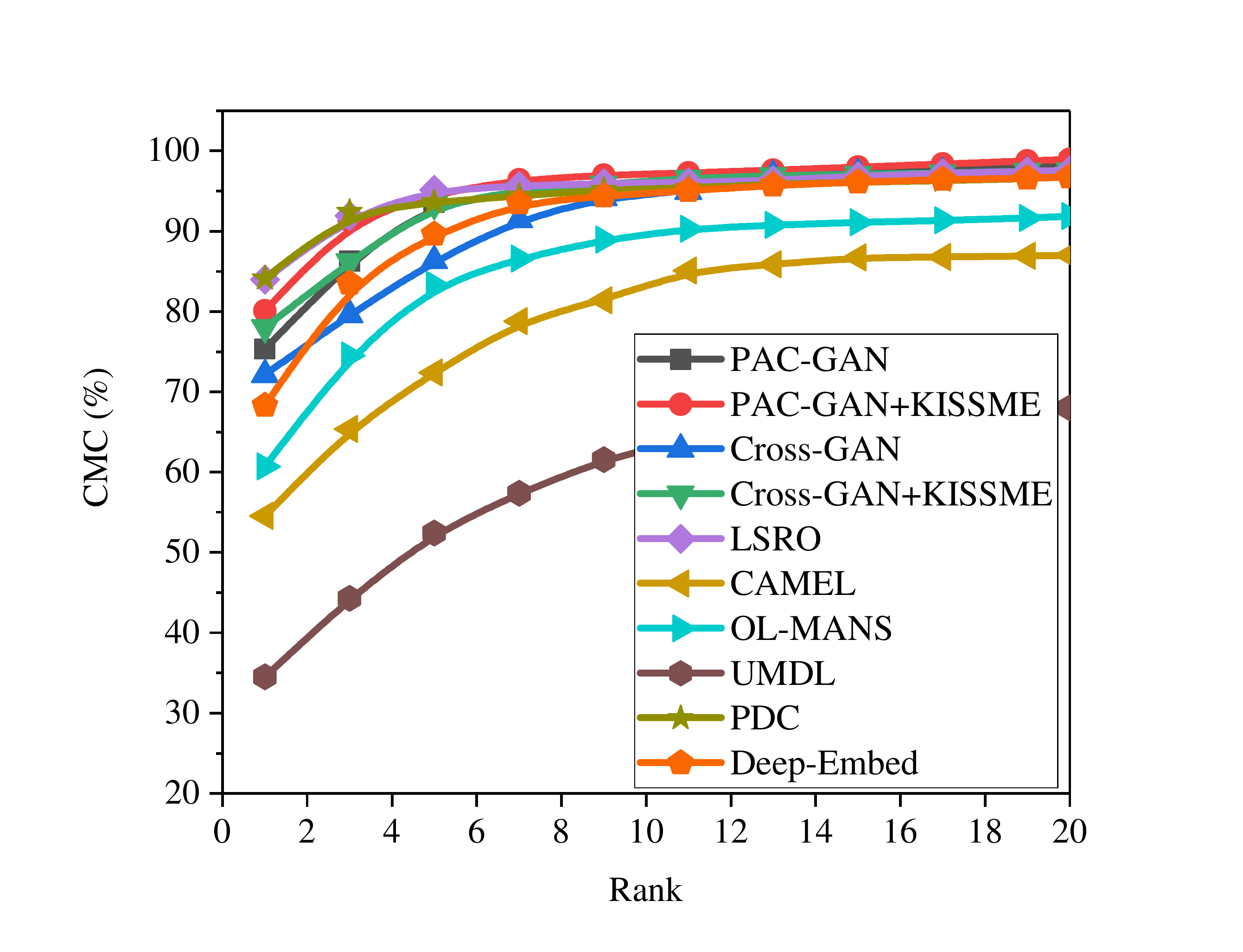}
         }
    \caption{The CMC curves of our approach and some state-of-the-art on three benchmarks: VIPeR, CUHK03 and Market-1501. To show these curves and their trends clearly, we just draw out the typical state-of-the-arts.}
    \label{fig:cmc-curve}
    \end{center}
    \end{minipage}
    \end{figure*}

The Rank-1, Rank-10, Rank-20 recognition rate of these methods on VIPeR, CUHK03 and Market-1501 are reported in Table~\ref{tab:results-VIPeR}, Table~\ref{tab:results-CUHK03} and Table~\ref{tab:results-Market-1501} respectively. The CMC curves of PAC-GAN, Cross-GAN and state-of-the-arts are illustrated in Fig.~\ref{fig:cmc-curve}. For the convenience of showing the comparion, we just draw out the typical state-of-the-arts, rather than all the methods aforementioned.

\subsubsection{Experiments on VIPeR dataset}

\begin{table}
    \centering
    \caption{The comparison results (Recognition rate: $R=1$, $R=10$, and $R=20$, test person$=316$) with state-of-the-arts on VIPeR dataset. The best performance values are bold-font.}
    \label{tab:results-VIPeR}
    \setlength{\tabcolsep}{3pt}
    \begin{tabular}{|p{110pt}|p{35pt}|p{35pt}|p{35pt}|}
    \hline
    Method (Semi/Un-supervised)&
    $R=1$&
    $R=10$&
    $R=20$\\

    \hline

    PAC-GAN&
    50.21&
    91.70&
    95.17\\

    PAC-GAN$+$KISSME~\cite{DBLP:conf/cvpr/KostingerHWRB12}&
    \textbf{56.33}&
    \textbf{95.15}&
    \textbf{98.04}\\

    Cross-GAN&
    49.28&
    91.66&
    93.47\\

    Cross-GAN$+$KISSME~\cite{DBLP:conf/cvpr/KostingerHWRB12}&
    54.25&
    94.30&
    97.05\\

    SDALF~\cite{DBLP:conf/cvpr/FarenzenaBPMC10}&
    19.87&
    49.37&
    65.73\\

    eSDC~\cite{Zhao_2013_CVPR}&
    26.31&
    58.86&
    72.77\\

    t-LRDC~\cite{DBLP:journals/pami/ZhengGX16}&
    27.40&
    46.00&
    75.10\\

    OSML~\cite{Bak_2017_CVPR}&
    34.30&
    -&
    -\\

    CAMEL~\cite{Yu_2017_ICCV}&
    30.90&
    52.00&
    72.50\\

    \hline
    Method (Supervised)&
    $R=1$&
    $R=10$&
    $R=20$\\
    \hline

    PDC~\cite{Su_2017_ICCV}&
    51.27&
    84.18&
    91.46\\

    Locally-Aligned~\cite{Li_2013_CVPR_Local}&
    29.60&
    69.30&
    86.70\\

    JointRe-id~\cite{Ahmed_2015_CVPR}&
    34.80&
    74.79&
    82.45\\

    SCSP~\cite{Chen2016Similarity}&
    53.54&
    91.49&
    96.65\\

    Multi-channel~\cite{Cheng_2016_CVPR}&
    47.80&
    84.80&
    91.10\\

    DNSL~\cite{Zhang_2016_CVPR}&
    42.28&
    82.94&
    92.06\\

    JSTL~\cite{Xiao_2016_CVPR}&
    38.40&
    -&
    -\\

    SI-CI~\cite{Wang_2016_CVPR}&
    35.80&
    83.50&
    -\\

    S-LSTM~\cite{DBLP:conf/eccv/VariorSLXW16}&
    42.40&
    79.40&
    -\\

    S-CNN~\cite{Varior2016Gated}&
    37.80&
    77.40&
    -\\

    SpindleNet~\cite{Zhao_2017_CVPR}&
    53.80&
    90.10&
    96.10\\

    Part-Aligned~\cite{Zhao_2017_ICCV}&
    48.70&
    87.70&
    93.00\\

    Deep-Embed~\cite{Lin2017Deep}&
    49.00&
    91.10&
    96.20\\

    LADF~\cite{Li_2013_CVPR}&
    29.34&
    75.98&
    88.10\\

    OL-MANS~\cite{Zhou2017Efficient}&
    44.90&
    74.40&
    93.60\\

    SalMatch~\cite{Zhao_2013_ICCV}&
    30.16&
    62.50&
    75.60\\

    DM$^3$~\cite{DBLP:journals/tcyb/0007H00JLS18}&
    37.52&
    80.85&
    88.90\\

    DeepList~\cite{DBLP:journals/tcsv/WangWGSH17}&
    40.36&
    81.20&
    91.08\\

    DDDM~\cite{DBLP:journals/tmm/WangHLYJYCL16}&
    22.35&
    66.08&
    76.32\\

    \hline
    \end{tabular}
    \label{tab1}
\end{table}
We evaluate our method PAC-GAN and Cross-GAN and state-of-the-arts in terms of CMC values, and the results are shown in Table~\ref{tab:results-VIPeR} and Fig.~\ref{fig:cmc-curve} (a). It is obvious that the proposed approach PAC-GAN has nice performance when $R=1$. The recognition rate of it is 50.21\%  that is higher than other unsupervised and semi-supervised methods. Besides, when $R=10$ and $R=20$, PAC-GAN achieves 91.70\% and 95.17\% respectively, which outperforms the semi/un-supervised mehtods including our previous method Cross-GAN (Rank-1 rate is 49.28\% , Rank-10 rate is 91.66\% and Rank-20 rate is 93.47\%) due to the performance boosting from pose augmentation.

Compared with the supervised state-of-the-arts, such as SpindleNet~\cite{Zhao_2017_CVPR} (Rank-20 rate is 96.10\%), Deep-Embed~\cite{Lin2017Deep} (Rank-20 rate is 96.20\%), the performance of our method is not the highest. However, PAC-GAN can still outdo the most competitors by attaining Rank-10 rate 91.70\% and Rank-20 95.17\%. It is particularly noteworthy that when we combine PAC-GAN and KISSME~\cite{DBLP:conf/cvpr/KostingerHWRB12} which is a supervised metric learning approach, the performance of recognition is improved substantially, higher than the matching rate of combination of Cross-GAN and KISSME. That means this novel unsupervised pose augmentation model can generate more effective visual representations for metric learning.

Fig.~\ref{fig:cmc-curve} (a) demonstrates the CMC curves of PAC-GAN and other semi/un-supervised and supervised state-of-the-arts on VIPeR dataset. Here we do not show all the approaches in Table~\ref{tab:results-VIPeR}. The matching rate of PAC-GAN and Cross-GAN increase gradually from Rank-1 to Rank-6 and then the growth is slowdown, which is much higher than SDALF~\cite{DBLP:conf/cvpr/FarenzenaBPMC10}, eSDC~\cite{Zhao_2013_CVPR}, SalMatch~\cite{Zhao_2013_ICCV} and CAMEL~\cite{Yu_2017_ICCV}. The trends of SpindleNet~\cite{Zhao_2017_CVPR} and Deep-Embed~\cite{Lin2017Deep} are very close to the proposed method, but they never surpass the combination of PAC-GAN and KISSME from Rank-1 to Rank-20.

\subsubsection{Experiments on CUHK03 dataset}

\begin{table}
    \centering
    \caption{The comparison results (Recognition rate: $R=1$, $R=10$, and $R=20$, test person$=100$) with state-of-the-arts on CUHK03 dataset. The best performance values are bold-font.}
    \label{tab:results-CUHK03}
    \setlength{\tabcolsep}{3pt}
    \begin{tabular}{|p{110pt}|p{35pt}|p{35pt}|p{35pt}|}
    \hline
    Method (Semi/Un-supervised)&
    $R=1$&
    $R=10$&
    $R=20$\\

    \hline

    PAC-GAN&
    85.66&
    97.71&
    99.54\\

    PAC-GAN$+$KISSME~\cite{DBLP:conf/cvpr/KostingerHWRB12}&
    \textbf{89.84}&
    98.98&
    99.75\\

    Cross-GAN&
    83.23&
    96.73&
    99.40\\

    Cross-GAN$+$KISSME~\cite{DBLP:conf/cvpr/KostingerHWRB12}&
    88.90&
    98.36&
    99.50\\

    OSML~\cite{Bak_2017_CVPR}&
    45.61&
    85.34&
    88.50\\

    LSRO~\cite{Zheng_2017_ICCV}&
    84.62&
    97.64&
    99.80\\

    eSDC~\cite{Zhao_2013_CVPR}&
    8.76&
    38.28&
    53.44\\

    CAMEL~\cite{Yu_2017_ICCV}&
    31.90&
    76.62&
    80.63\\

    UMDL~\cite{Peng_2016_CVPR}&
    1.64&
    8.34&
    10.24\\

    \hline
    Method (Supervised)&
    $R=1$&
    $R=10$&
    $R=20$\\
    \hline

    MSCAN~\cite{Li_2017_CVPR}&
    74.21&
    97.54&
    99.25\\

    SSM~\cite{Bai_2017_CVPR}&
    71.82&
    92.54&
    96.64\\

    PDC~\cite{Su_2017_ICCV}&
    88.70&
    \textbf{99.24}&
    99.67\\

    DNSL~\cite{Zhang_2016_CVPR}&
    58.90&
    92.45&
    96.30\\

    JointRe-id~\cite{Ahmed_2015_CVPR}&
    54.74&
    91.50&
    97.31\\

    E-Metric~\cite{DBLP:conf/eccv/ShiYZLLZL16}&
    61.32&
    96.50&
    97.50\\

    S-LSTM~\cite{DBLP:conf/eccv/VariorSLXW16}&
    57.30&
    88.30&
    -\\

    S-CNN~\cite{Varior2016Gated}&
    61.80&
    88.30&
    -\\

    Deep-Embed~\cite{Lin2017Deep}&
    73.00&
    94.60&
    98.60\\

    SpindleNet~\cite{Zhao_2017_CVPR}&
    88.50&
    98.80&
    99.20\\

    Part-Aligned~\cite{Zhao_2017_ICCV}&
    85.40&
    98.60&
    \textbf{99.90}\\

    XQDA~\cite{Liao2015Person}&
    52.20&
    92.14&
    96.25\\

    OL-MANS~\cite{Zhou2017Efficient}&
    61.70&
    92.40&
    98.52\\

    DM$^3$~\cite{DBLP:journals/tcyb/0007H00JLS18}&
    56.16&
    91.31&
    96.74\\

    DeepList~\cite{DBLP:journals/tcsv/WangWGSH17}&
    54.84&
    92.56&
    96.61\\

    DDDM~\cite{DBLP:journals/tmm/WangHLYJYCL16}&
    19.58&
    49.64&
    63.25\\

    \hline
    \end{tabular}
    \label{tab1}
\end{table}
Table~\ref{tab:results-CUHK03} shows the matching rates of these approaches on CUHK03 dataset. The Rank-1 matching rate of PAC-GAN is higher than all these semi/un-supervised methods, and it just lower than two supervised state-of-the-arts, namely PDC~\cite{Su_2017_ICCV} (Rank-1=88.70\%) and SpindleNet~\cite{Zhao_2017_CVPR} (Rank-1=88.50\%). Like the situation on VIPeR dataset, the performance can be boosted obviously by using KISSME~\cite{DBLP:conf/cvpr/KostingerHWRB12} as auxiliary  enhancement with PAC-GAN. That is, the recognition rate of this combination is improved from Rank-1=85.66\% to 89.84\%. The Rank-10 and Rank-20 matching rates of PAC-GAN combing with KISSME are 98.98\% and 99.54\%, which is higher than most of these methods and just a little less than PDC~\cite{Su_2017_ICCV} (Rank-10=99.24\%) and Part-Aligned~\cite{Zhao_2017_ICCV} (Rank-20=99.90\%), respectively. However, PAC-GAN do not need any labeled data for training, which is a major benefit advantage for common person Re-Id task.

The CMC curves of the proposed method and the competitors on CUHK03 dataset are shown in Fig.~\ref{fig:cmc-curve} (b). The trends of PAC-GAN, Cross-GAN, SpindleNet~\cite{Zhao_2017_CVPR}, and LSRO~\cite{Zheng_2017_ICCV} are very similar, which grow step-By-step from 84\% around to about 96\% in the interval of Rank-1 to Rank-6. After that the change of them tend to be gentle and the peak CMC values of them are very close, much higher than OSML~\cite{Bak_2017_CVPR} amd CAMEL~\cite{Yu_2017_ICCV}.

\subsubsection{Experiments on Market-1501 dataset}
The comparision results on Market-1501 dataset reported in Table~\ref{tab:results-Market-1501} demonstrates that the matching rate of the proposed method (Rank-1=75.34, Rank-10=95.71, Rank-20=98.45) is higher than Cross-GAN (Rank-1=72.15, Rank-10=94.31, Rank-20=97.50) as the more priors are provided by pose augmentation. However, the performance of PAC-GAN is lower than LSRO~\cite{Zheng_2017_ICCV} when $R=1$ because on one hand many pedestrian samples in Market-1501 dataset have similar appearance, and on the other hand LSRO can produce more authentic pedestrian samples for discrimination. It is undeniable that the computation for a great quantity of more authentic samples generation is very expensive. By contrast, our method has obvious advantage that PAC-GAN achieves good performance via pose augmentation but it do not require any labeled data for training. Similar to the above-mentioned results, with the assistance of supervised metric learning technique, KISSME~\cite{DBLP:conf/cvpr/KostingerHWRB12}, the combined solution (PAC-GAN+KISSME) can ourperform all these semi/un-supervised and supervised approaches with higher matching rates Rank-10=97.02\% and Rank-20=98.94\%.

The CNC curves of PAC-GAN and other methods on Market-1501 dataset are drawn in Fig.~\ref{fig:cmc-curve} (c). Like the evaluation on CUHK03, the growths of LSRO~\cite{Zheng_2017_ICCV}, PAC-GAN, Cross-GAN and their combination solutoins are very close, which are fast in the interval Rank-1 to Rank-5 and gradually slowdown with rank increasement. No doubt, in Market-1501 dataset the performance of the proposed method is on the whole the best among them.

\begin{table}
    \centering
    \caption{The comparison results (Recognition rate: $R=1$, $R=10$, and $R=20$, test person$=751$) with state-of-the-arts on Market-1501 dataset. The best performance values are bold-font.}
    \label{tab:results-Market-1501}
    \setlength{\tabcolsep}{3pt}
    \begin{tabular}{|p{110pt}|p{35pt}|p{35pt}|p{35pt}|}
    \hline
    Method (Semi/Un-supervised)&
    $R=1$&
    $R=10$&
    $R=20$\\

    \hline

    PAC-GAN&
    75.34&
    95.71&
    98.45\\

    PAC-GAN$+$KISSME~\cite{DBLP:conf/cvpr/KostingerHWRB12}&
    80.06&
    \textbf{97.02}&
    \textbf{98.94}\\

    Cross-GAN&
    72.15&
    94.31&
    97.50\\

    Cross-GAN$+$KISSME~\cite{DBLP:conf/cvpr/KostingerHWRB12}&
    78.03&
    96.25&
    97.50\\

    eSDC~\cite{Zhao_2013_CVPR}&
    33.45&
    60.61&
    67.53\\

    SDALF~\cite{DBLP:conf/cvpr/FarenzenaBPMC10}&
    20.53&
    -&
    -\\

    LSRO~\cite{Zheng_2017_ICCV}&
    83.97&
    95.64&
    97.56\\

    CAMEL~\cite{Yu_2017_ICCV}&
    54.56&
    84.67&
    87.03\\

    PUL~\cite{DBLP:journals/tomccap/FanZYY18}&
    45.53&
    72.75&
    72.65\\

    UMDL~\cite{Peng_2016_CVPR}&
    34.54&
    62.60&
    68.03\\

    BoW~\cite{DBLP:conf/iccv/ZhengSTWWT15}&
    34.40&
    -&
    -\\

    \hline
    Method (Supervised)&
    $R=1$&
    $R=10$&
    $R=20$\\
    \hline

    JSTL~\cite{Xiao_2016_CVPR}&
    44.72&
    77.24&
    82.00\\

    SSM~\cite{Bai_2017_CVPR}&
    82.21&
    -&
    -\\

    CADL~\cite{Lin_2017_CVPR}&
    73.84&
    -&
    -\\

    PDC~\cite{Su_2017_ICCV}&
    \textbf{84.14}&
    94.92&
    96.82\\

    MSCAN~\cite{Li_2017_CVPR}&
    80.31&
    -&
    -\\

    SCSP~\cite{Chen2016Similarity}&
    51.90&
    -&
    -\\

    DNSL~\cite{Zhang_2016_CVPR}&
    61.02&
    -&
    -\\

    S-CNN~\cite{Varior2016Gated}&
    65.88&
    -&
    -\\

    Deep-Embed~\cite{Lin2017Deep}&
    68.32&
    94.59&
    96.71\\

    SpindleNet~\cite{Zhao_2017_CVPR}&
    76.90&
    -&
    -\\

    XQDA~\cite{Liao2015Person}&
    43.79&
    75.32&
    80.41\\

    OL-MANS~\cite{Zhou2017Efficient}&
    60.72&
    89.80&
    91.87\\

    DM$^3$~\cite{DBLP:journals/tcyb/0007H00JLS18}&
    72.26&
    90.67&
    94.10\\

    DeepList~\cite{DBLP:journals/tcsv/WangWGSH17}&
    71.39&
    89.40&
    94.55\\

    DDDM~\cite{DBLP:journals/tmm/WangHLYJYCL16}&
    74.42&
    93.75&
    95.20\\

    \hline
    \end{tabular}
    \label{tab1}
\end{table}

\section{Conclusion}\label{sec:conclusion}
In this paper we propose to enhance the performance of unsupervised cross-view person Re-Id by introducing a novel pose augmentation cross-view person Re-Id scheme called PAC-GAN. In this scheme, a novel deep generative model named CPG-Net is developed to produce new samples that have various poses from skeleton samples and original pedestrian samples. A pose augmented dataset is generated by combing the new samples and original samples, which are fed into the person Re-Id model named Cross-GAN to improve the identification accuracy. The results of our experiments illustrate that the performance of unsupervised cross-view person Re-Id can be improved obviously by using PAC-GAN.

\section*{Acknowledgment}
This work was supported in part by the National Natural Science Foundation of China (61702560, 61836016, 616721\\77), project (2018JJ3691, 2016JC2011) of Science and Technology Plan of Hunan Province, and the Research and Innovation Project of Central South University Graduate Students(2018zzts177, 2018zzts588).

\bibliographystyle{spmpsci}      % mathematics and physical sciences
\bibliography{allbib}

\end{document}